\documentclass[sn-basic,Numbered]{sn-jnl}

\usepackage{graphicx}%
\usepackage{multirow}%
\usepackage{amsmath,amssymb,amsfonts}%
\usepackage{amsthm}%
\usepackage{mathrsfs}%
\usepackage[title]{appendix}%
\usepackage{xcolor}%
\usepackage{textcomp}%
\usepackage{manyfoot}%
\usepackage{booktabs}%
\usepackage{algorithm}%
\usepackage{algorithmicx}%
\usepackage{algpseudocode}%
\usepackage{listings}%
\usepackage{array}
\usepackage{multicol, makecell}
\usepackage{diagbox}
\usepackage{subcaption}

\newcommand{\dquotes}[1]{``#1''}

\makeatletter
\newcommand\footnoteref[1]{\protected@xdef\@thefnmark{\ref{#1}}\@footnotemark}
\makeatother

\usepackage{color}

\begin{document}

\title[Article Title]{Improving Neural Radiance Fields Using Near-Surface Sampling With Point Cloud Generation}


\author[1]{\fnm{Hye Bin} \sur{Yoo}}\email{gpqls7662@g.skku.edu}
\equalcont{These authors contributed equally to this work.}

\author[2]{\fnm{Hyun Min} \sur{Han}}\email{manduss1204@gmail.com}
\equalcont{These authors contributed equally to this work.}

\author*[2]{\fnm{Sung Soo} \sur{Hwang}}\email{sshwang@handong.edu}

\author*[1,3]{\fnm{Il Yong} \sur{Chun}}\email{iychun@skku.edu}

\affil[1]{\orgdiv{Department of Electrical and Computer Engineering}, \orgname{Sungkyunkwan University}, \orgaddress{\city{Suwon}, \postcode{16419}, \country{Republic of Korea}}}

\affil[2]{\orgdiv{Department of Information and Communication Engineering}, \orgname{Handong Global University}, \orgaddress{\city{Pohang}, \postcode{37554}, \country{Republic of Korea}}}

\affil[3]{\orgdiv{Departments of Artificial Intelligence, Semiconductor Convergence Engineering, and Display Convergence Engineering,
and
Center for Neuroscience Imaging Research, Institute for Basic Science}, \orgname{Sungkyunkwan University}, \orgaddress{\city{Suwon}, \postcode{16419}, \country{Republic of Korea}}}


\abstract{Neural radiance field (NeRF) is an emerging view synthesis method that samples points in a three-dimensional (3D) space and estimates their existence and color probabilities.
The disadvantage of NeRF is that it requires a long training time since it samples many 3D points. 
In addition, if one samples points from occluded regions or in the space where an object is unlikely to exist,
the rendering quality of NeRF can be degraded.
These issues can be solved by estimating the geometry of 3D scene.  
This paper proposes a near-surface sampling framework to improve the rendering quality of NeRF.
To this end, the proposed method estimates the surface of a 3D object using depth images of the training set and performs sampling only near the estimated surface.
To obtain depth information on a novel view, the paper proposes a 3D point cloud generation method and a simple refining method for projected depth from a point cloud. 
Experimental results show that the proposed near-surface sampling NeRF framework can significantly improve the rendering quality,
compared to the original NeRF and three different state-of-the-art NeRF methods.
In addition, one can significantly accelerate the training time of a NeRF model with the proposed near-surface sampling framework.}




\maketitle

\section{Introduction}

Recently, metaverse and virtual reality applications are rapidly drawing attention. 
In such applications, it is important to generate novel views accurately. 
One way to achieve this goal is to generate a three-dimensional (3D) model first and follow a conventional rendering pipeline \cite{3D}. 
However, generating a 3D model needs a lot of time and effort.

Image-based rendering (IBR) is another approach that generates novel views without explicitly generating a 3D model. 
Several methods generate a novel view using image morphing \cite{VI}. 
The Layered Depth Images method \cite{LDI} stores multiple depth and color values for each pixel to effectively fill the hole behind the foreground object in a novel view.
Light fields \cite{LF} and Lumigraph \cite{Lumigraph} that express light rays as a function were also proposed.

Recently, among IBR methods, neural radiance field (NeRF) \cite{NeRF} has been rapidly gaining attention.
Ray, a core concept of NeRF, means lines shot in a straight line from the camera position to an object.
A NeRF network predicts the color and density of each point utilizing 3D points sampled from each ray.
Then a novel view is obtained by performing a line integral using this color and density.

\begin{figure}
\centering
\includegraphics[width=\textwidth]{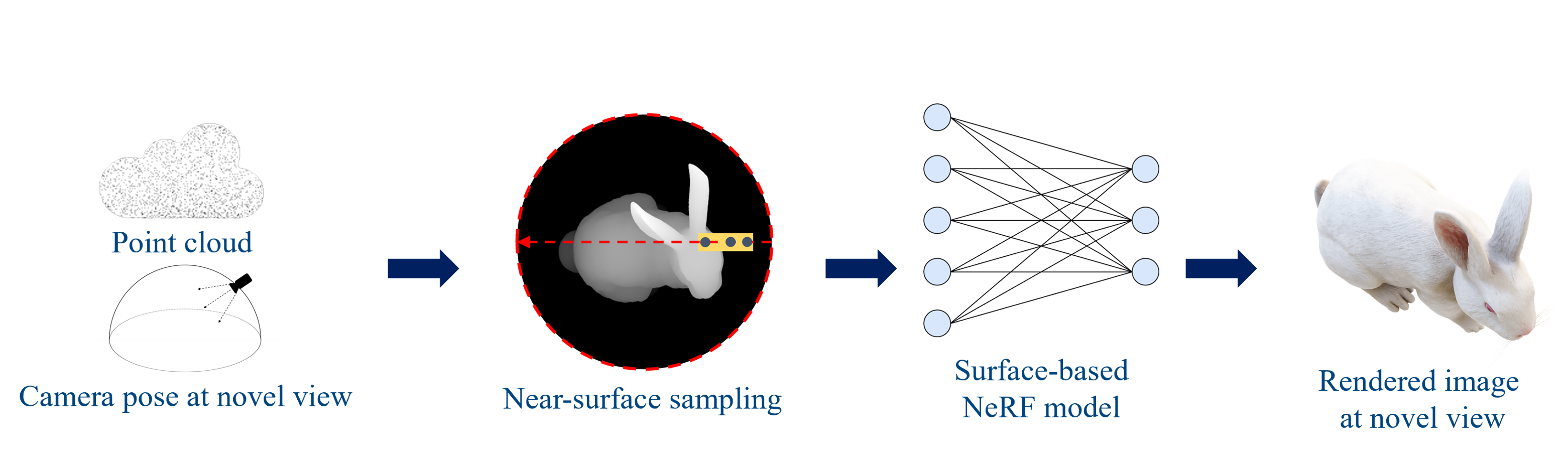}
\vspace{-0.5pc}
\caption{The brief overview of the proposed NeRF framework that samples points near the estimated surface from a point cloud.}
\label{simplify_process}
\end{figure}

The original NeRF \cite{NeRF} performs sampling within a range that includes the entire 3D object.
This paper proposes to use depth information to sample 3D points only around surface of an object in NeRF,
where we consider the practical scenario that depth information is only available at hands (from depth cameras) in a training dataset.
To consider that measured/estimated depths maps may be inaccurate due to capturing environments, 
we propose to generate a 3D point cloud using available (inaccurate) depth information in training,
and to use this 3D point cloud to estimate a depth image for each novel view in test (i.e., inference).
Figure~\ref{simplify_process} illustrates the brief overview of the proposed NeRF framework.
Simply projecting a 3D point cloud onto a novel view generates a rather rough depth image.
To obtain more accurate depth images, we additionally propose a refining method that removes unnecessary 3D points in generating a point cloud and fills the hole of the projected depth image.
Simply put, to improve NeRF, the paper proposes an advanced sampling method around the surface of an object/a scene using estimated depth images from generated point cloud.
Our experimental results with different datasets demonstrate that the proposed framework outperforms
original NeRF and three different state-of-the-art NeRF methods.

The rest of the paper is organized as follow.
Section~2 reviews NeRF and its follow-up works with particularly related works with ours, and presents differences between the proposed NeRF and existing depth-based NeRFs.
Section~3 provides motivation and detail of the proposed method,
Section~4 reports experiments and analysis, and
Section~5 discusses conclusions, limitation and future work.

\section{Related works}

\subsection{NeRF}
\label{sec:nerf}

NeRF \cite{NeRF} is a state-of-the-art view synthesis technology that samples points on rays and synthesizes views through differentiable volume rendering. 
The input of this algorithm is a single continuous five-dimensional (5D) coordinate consisting of a 3D spatial location and a two-dimensional viewing direction.
The output is a volume density and view-dependent emitted radiance at the corresponding spatial location.
In other words, 
the key idea of NeRF is to train a neural network that predicts a view-dependent color value and a volume probability value by taking a 5D coordinate.
Using those two predicted values, a final rendered color value is determined by performing a line integral with classical volume rendering. 
To further improve the rendering quality, NeRF uses the following two techniques: positional encoding and hierarchical volume sampling.
Positional encoding increases the dimension of input data; the hierarchical volume sampling technique allocates more samples to regions that are expected to include visible content.
Hierarchical volume sampling is named as it performs sampling with two different networks, \dquotes{coarse} one and \dquotes{fine} one.
For each ray, a coarse network gives a view-dependent emitted color and volume density using $N_\text{c}$ points that are sampled with stratified sampling method along the ray.
A piecewise-constant probability density function (PDF) is generated (along each ray) by normalizing contribution weights that are calculated with volume densities and the distances between adjacent samples of $N_\text{c}$ points.
After integrating the generated PDF to calculate cumulative distribution function, $N_\text{f}$ points are sampled through inverse transform sampling.
A fine network gives a view-dependent color value and volume density using $N_\text{c}$ points and those more informed $N_\text{f}$ points.
After all, one calculates the final rendering of the corresponding ray with $N_\text{c} + N_\text{f}$ points.
Through this process, NeRF can represent a 3D object (in 360 degrees) and forward-facing scenes with continuous views.
However, NeRF in its original form has several limitations.
For example, it can represent only static scenes;
its training and inference is slow;
one NeRF network represents only one object/scene.

\subsection{Follow-up works of NeRF}

Researchers has been improving the original NeRF model \cite{NeRF} in various aspects.
The first aspect is to reduce training time of NeRF models while maintaining rendering accuracy \cite{efficientnerf, Depth-supervised-NeRF, Point-NeRF, instant-ngp}.
\cite{efficientnerf}
reduces training time by proposing a new sampling method to use less number of samples per ray.
\cite{Depth-supervised-NeRF} 
supervises depth to use a smaller number of views in training.
\cite{Point-NeRF} can accelerate training by quickly generating an initial rough point cloud and refining it in an iterative manner.
\cite{instant-ngp} uses a learnable encoding method instead of positional encoding, and update only parameters related to sampling positions instead of updating all parameters.

The second aspect is to improve inference time of NeRF models \cite{Derf, voxel_fields, AutoInt, DONeRF, instant-ngp}.
\cite{Derf} and \cite{voxel_fields} reduces inference time by spatially decomposing and processing the scene:
\cite{Derf} uses a spatially decomposed scene and a small network for each space; 
\cite{voxel_fields} skips spaces with irrelevant scenes among the decomposed spaces during inference.
\cite{AutoInt} uses volume integral calculation network instead of the classical integral calculation method to shorten inference.
\cite{DONeRF} uses a rendering pipeline that includes a network to predict the optimal sample locations on rays to reduce inference time.
Using learnable encoding method instead of positional encoding 
\cite{instant-ngp} can accelerate inference.

Third aspect is to consider different scenarios with NeRF models \cite{INeRF, BARF, Fig-NeRF, Giraffe, GeoNeRF, Nerfies, D-nerf, Nerv, Nerd, NeRF_wild}.
\cite{INeRF} additionally estimates camera pose.
\cite{BARF} considers the case that camera poses are imperfect or unknown.
\cite{Fig-NeRF, Giraffe, GeoNeRF} consider multi-object/scene representation.
In particular, 
\cite{Fig-NeRF} disentangles foreground and background.
Dynamic scene representation \cite{Nerfies, D-nerf} and relighting \cite{Nerv, Nerd, NeRF_wild} makes NeRF to be applicable to changing scenes rather than static scenes.

\subsection{Depth-based NeRFs and their relations with the proposed NeRF framework}

Depth oracle neural radiance field (DONeRF) \cite{DONeRF} uses ground-truth depth images of the training set to train ideal sample locations on rays, and performs sampling in the estimated locations. 
However, DONeRF works only on forward-facing scenes where all camera poses belong to a bounding box called the view cell.
Depth supervised neural radiance field (DSNeRF) \cite{Depth-supervised-NeRF} uses a sparse depth map estimated with the structure from motion technique and adds an optimization process to the original NeRF using estimated depth information, to achieve the best rendering performance of original NeRF with fewer training iterations and images.

Similar to DONeRF, we aim to improve the quality of rendered images by using depth images available at hand in a training dataset.
Note, however, that different from DONeRF, the proposed method does not use the view cell information that is required in DONeRF, and is applicable with less restricted camera positions.
Similar to DSNeRF, we use depth information by leveraging a point cloud.
However, the proposed framework and DSNeRF use a point cloud in a different way.
DSNeRF uses a point cloud to adjust the volume density function of NeRF.
Different from this, the proposed framework uses a point cloud to directly estimate the distance to the object surface from a camera.

\section{Proposed method}

\subsection{Motivation}

\begin{figure}
\centering
\includegraphics[width=0.55\linewidth]{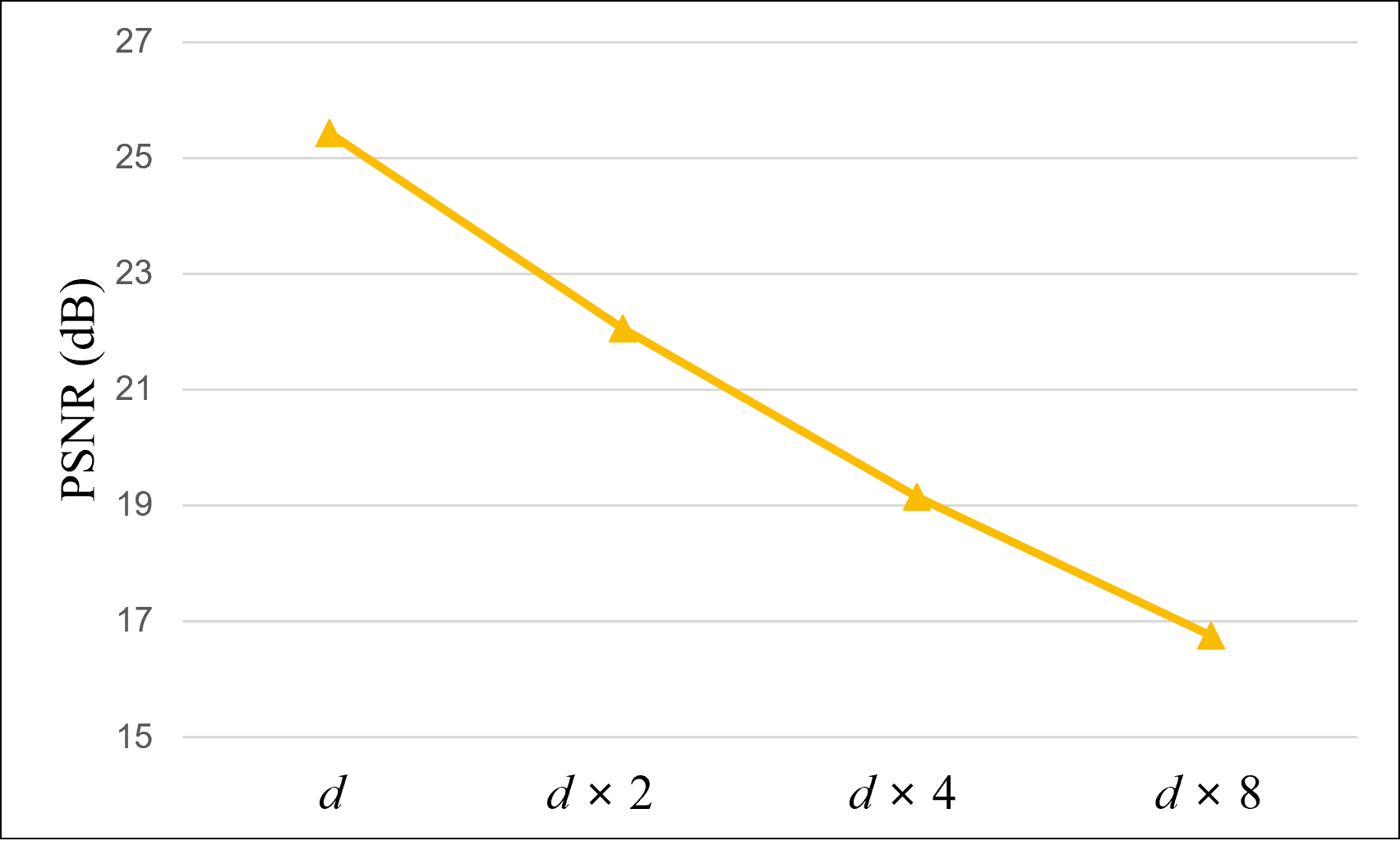}
\caption{The NeRF rendering accuracy comparisons with different sampling ranges. Here, $d$ denotes the default sampling range of NeRF.}
\label{fig1}
\end{figure} 

In NeRF \cite{NeRF}, there exists a room for improvement of rendering accuracy.
NeRF uses a hierarchical volume sampling method that performs sampling twice:
\dquotes{rough} sampling with a stratified sampling approach and \dquotes{fine} sampling in the space where an object is likely to exist.
See details in Section~\ref{sec:nerf}.
The stratified sampling approach in NeRF divides a specified range many bins and selects a sample uniformly a random from each bin.
In the stratified sampling process, sampling is performed not only in the space where the object exists, but also in the free space or the occluded region. 
Sampling in free space and occluded region may degrade rendering quality.
If one can sample points only around an object in the rough sampling stage, the rendering performance might improve even without the fine sampling process.

To show the effects of the sampling density around an object on the rendering quality, we ran simple experiments with different sampling ranges around the surface of an object.  
Figure \ref{fig1} shows the rendering accuracy with peak signal-to-noise ratio (PSNR) values with different sampling range, where we increased the default sampling range of NeRF by a factor of $2$, $4$, and $8$ by increasing distances between two samples.
As the sampling range increases, i.e., sampling density around an object decreases, the rendering accuracy rapidly degrades. 
We observed from theses experiments that narrowing the sampling range around an object can improve the rendering quality in NeRF.
This corresponds to the hierarchical volume sampling scheme of original NeRF that re-extracts samples with high volume density values to increase rendering efficiency.

Recently, diverse low-cost depth cameras with high accuracy have been proposed \cite{kinect, realsense}.
Depth cameras (using multi-view) can measure the distance between an object and the device, giving additional 3D information of an object.
We conjecture that if we sample points on 3D ray only around the surface of an object, the rendering quality of NeRF improves.

\subsection{Overview}

\begin{figure*}
\centering
\includegraphics[width=1.02\linewidth]{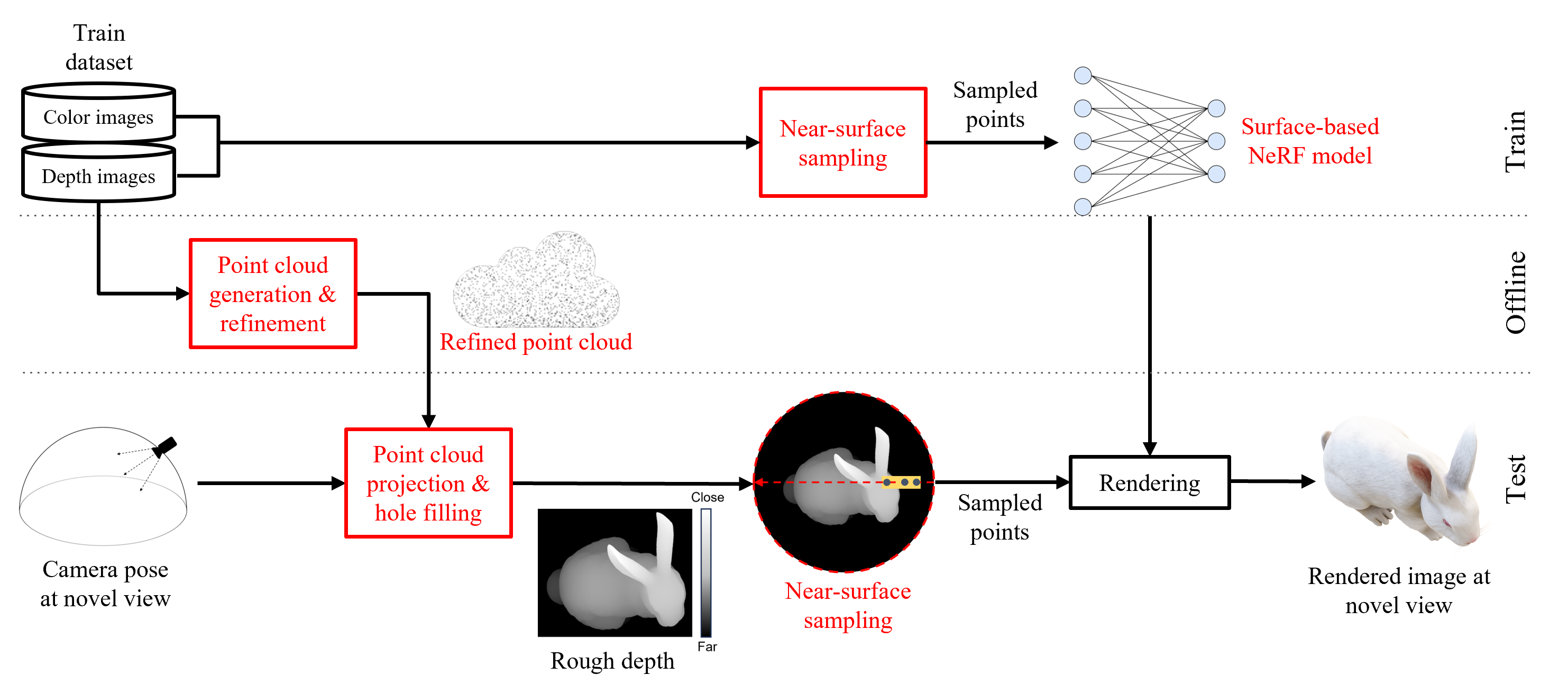}
\caption{The overall diagram of the proposed NeRF framework.
The red words highlight proposed modules. 
}
\label{fig2}
\end{figure*}  

Figure~\ref{fig2} illustrates the overall process of the proposed framework. 
A training set consists of color images and depth images, and at the train stage we use both.
In particular, we use depth images to sample in the area close to the surface of the object in a 3D space, and we refer this sampling strategy as surface-based sampling.
By using those sample points obtained through surface-based sampling, we train the NeRF model.
At the offline stage,
we use depth images of the training set to generate a point cloud and save this point cloud for inference.
At the test stage,
we use the saved point cloud at the offline stage to generate a depth image corresponding to a novel view.
We further refine depth images through computationally efficient hole filling for surface-based sampling.
Using sampled points only around the surface of an object that is estimated with a refined depth, 
we render images of novel views with a single NeRF network.

\subsection{Surface-based sampling}
\label{sec:surface-sample}

\begin{figure}[t!]
\centering
\includegraphics[width=7.5cm]{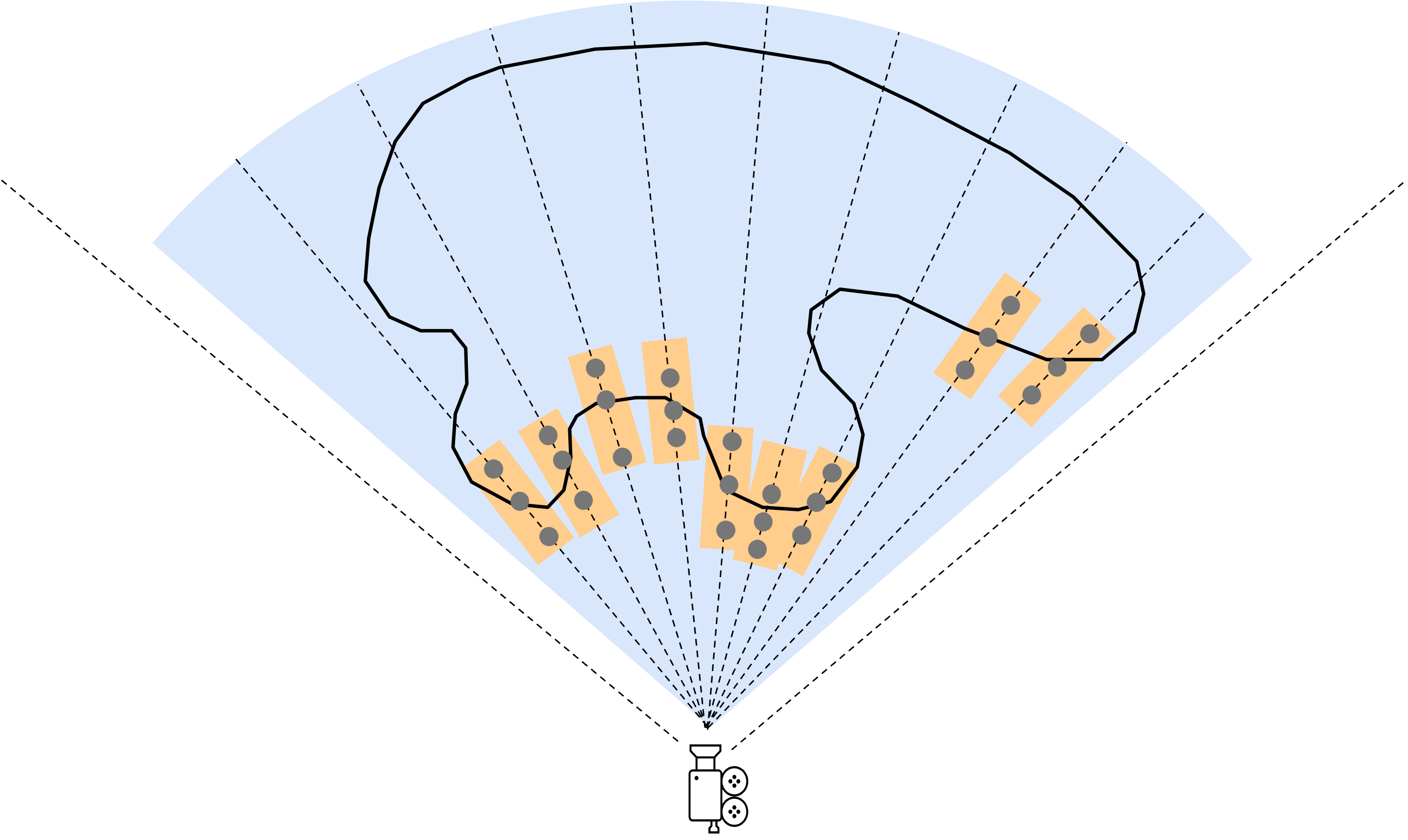}

\caption{Sampling range comparisons between the original NeRF (blue) and the proposed surface-based sampling scheme (orange).
The solid line represents the surface of an object, and the dotted lines inside the blue fan represent rays. 
The area within two dotted lines outside the blue region corresponds to the field of view of a camera.
Different the original NeRF, the proposed method samples only around the surface of an object.
}
\label{fig3}
\end{figure}   

Figure~\ref{fig3} illustrates the difference between the sampling range of the original NeRF's sampling method (blue) and that of surface-based sampling method (orange).
Different from original NeRF that samples 3D points at a wide range that includes the entire 3D object, the proposed surface-based sampling method mainly samples those around the surface of the object.

We now describe the geometry of the proposed surface-based sampling method for each ray of each view.
As in the original NeRF, we assume that each ray is propagated from the location of a camera (see Figure~\ref{fig3}).
We define the location of a camera in each ray as $0$.
The distance between the locations of a camera and an object is the depth value from a depth image, and we denote it as $d$.
Let the half of some specified sampling range be $\alpha$.
Then, the location of a point nearest to the camera within the sampling range can be calculated as follows:
\begin{equation}\label{eq1}
S_{r,0} = d - \alpha,
\end{equation}
Now, we determine the location of the $n$th sample for each ray (considering that a ray is originated from the camera location, $0$) by
\begin{equation}
\label{eq2}
S_{r, n} = S_{r, 0} + (n-1)\frac{2\alpha}{N} + \gamma, \quad n = 1,\ldots,N,
\end{equation}
where $N$ is the number of sample points for each ray, and $\gamma$ is a random number generated between $0$ and $2\alpha/N$.
We perform stratified sampling near the surface of an object, where we determine the sample locations by (\ref{eq2}).
In (\ref{eq2}), $[0, 2\alpha/N]$ is the length of each bin in stratified sampling of the original NeRF method.
Here, the parameter $\alpha$ determines the sampling range; if $N$ is fixed, $\alpha$ ultimately affects the sampling density around the surface.
As $\alpha$ decreases, the length of each bin is shorter and distances between sample points are expected to become close, so the sampling density near the surface increases.
As $\alpha$ increases, the length of each bin is longer and distances between sample points are expected to become far, so the sampling density near the surface decreases.

Different from the two-step network sampling scheme in original NeRF, 
the proposed framework directly samples points near the surface of an object by using depth information in the near-surface sampling scheme (\ref{eq2}) in a single step, i.e., it uses a single network.
We expect that if the depth to the surface of a 3D object $d$ is accurately estimated, 
the rendering quality improves by using small $\alpha$, i.e., densely sampling 3D points.
If it is poorly estimated, we expect that small $\alpha$ rather degrades the rendering quality. 
With fixed $N$, we recommend setting $\alpha$ considering the accuracy of depth images.

\begin{figure*}
\centering
\includegraphics[width=\linewidth]{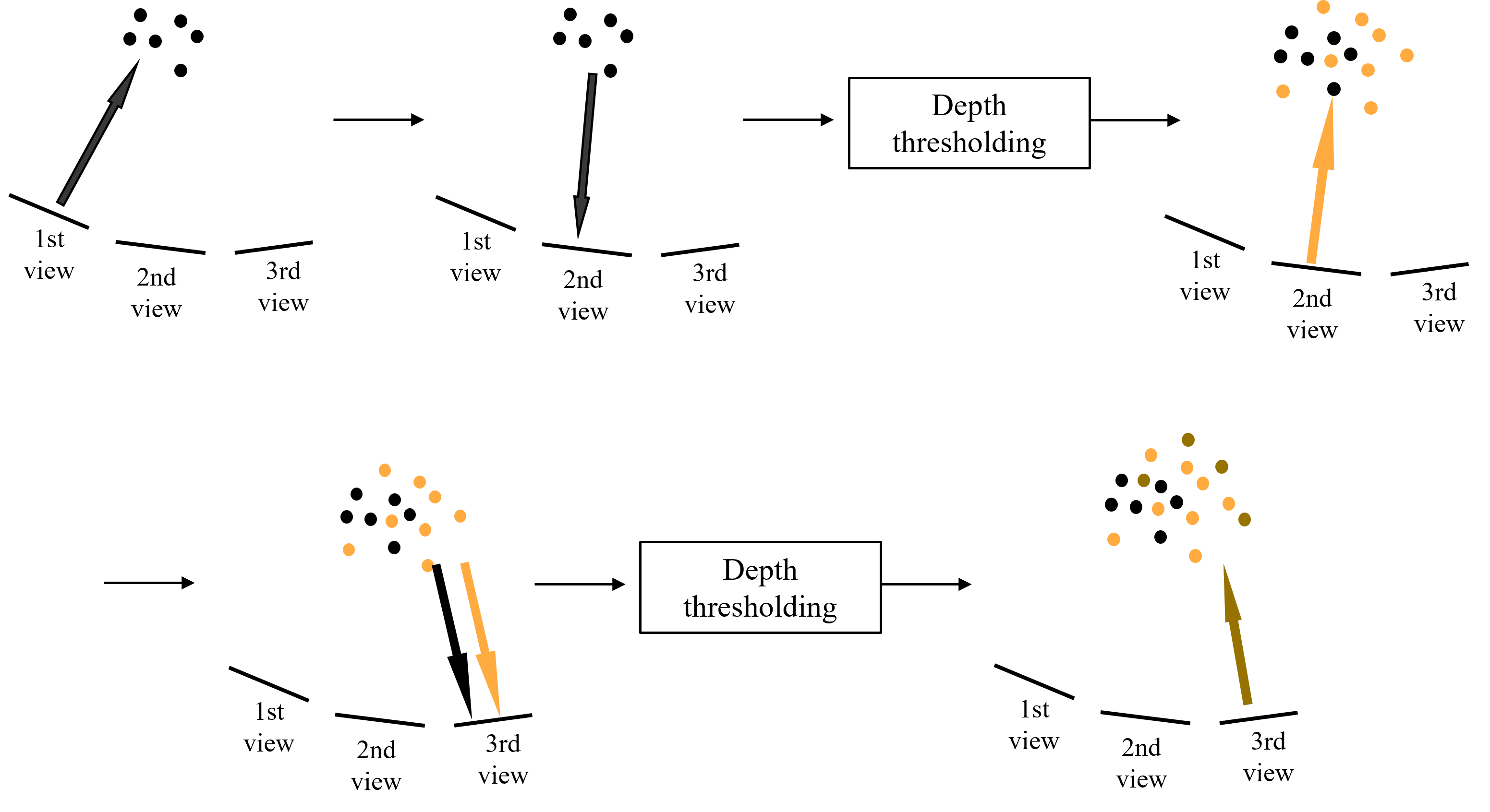}

\caption{An example of the proposed point cloud refinement.
In the first step, we generate a point cloud from a depth image of a viewpoint.
In the second step, 
we project the generated point cloud onto the next viewpoint.
In the third step, 
we use the depth thresholding scheme (\ref{eq3}) using projected points in the next viewpoint and ground-truth depth values.
If a projected point in the next viewpoint has a similar value to the ground-truth, we consider that the corresponding 3D point is redundant to generate.
We then generate new 3D points in the next viewpoint if they are determined to be necessary.
We repeat the above steps.}
\label{PCprocess}
\end{figure*}

\subsection{Depth image generation for novel views}

In the training stage, we perform surface-based sampling without any additional process, assuming that a depth image for each view is available.
In the test stage, however, we assume that depth images are \emph{unavailable,} so we perform depth estimation for a novel view for surface-based sampling. 
For depth estimation, in the offline stage, we generate and save a point cloud as shown in Figure~\ref{fig2}.
In the test stage, we use this point cloud to estimate depth images for novel views.
Using this depth estimation process, surface-based sampling can be performed without a ground truth depth image in the test stage.

\subsubsection{Point cloud generation and refinement in the offline stage}
\label{sec:point-gen}

Figure \ref{PCprocess} illustrates the key concept of the proposed point cloud generation and refinement method. 
To improve the accuracy of depth estimation, we generate 3D points with a subset of training images, by repeatedly eliminating inaccurate points.
In constructing a subset of training images, we 
give a sufficient and uniform distance between their adjacent viewpoints. 
This setup is more efficient in constructing a 3D point cloud,
compared to the setup that uses the entire training views.
See details of this experimental setup later in Section~\ref{sec:experimental setup}.

Each iteration consists of the following four steps and we repeat them with the cardinality of a subset of training images, where we sequentially follow the trajectory of viewpoints in a subset of training data:
\begin{enumerate}
\item[\textit{1)}] We generate a point cloud using a depth image from a viewpoint.

\item[\textit{2)}] We project 3D points of the generated point cloud onto an image plane of the next viewpoint, and obtain the distance between each 3D point and the camera location of the next viewpoint by using the multiple view geometry calculation method \cite{MVG}.

\item[\textit{3)}] We compare each calculated distance to a ground-truth depth value from the depth image at the next viewpoint, and identify if the following condition is satisfied:
\begin{equation}
\label{eq3}
\big| \tilde{d} - d_{\text{GT}} \big| \leq \tau,
\end{equation}
where $\tilde{d}$ denotes the calculated distance using the second step above, and
$d_{\text{GT}}$ denotes the ground-truth depth value of a pixel position where the 3D point is projected, and $\tau$ denotes some specified threshold.

\item[\textit{4)}] If the condition (\ref{eq3}) is \emph{not} satisfied, we generate a new 3D point by back-projecting a pixel of the value $d_{\text{GT}}$.
\end{enumerate}
Setting $\tau$ appropriately is important to generate an accurate point cloud. 
If $\tau$ is too large, 3D points with similar locations will be considered as the same point.
Consequently, fewer 3D points are generated, leading to faster rendering times; 
however, estimated depth images may contain many holes.
Conversely, if $\tau$ is too small, the number of 3D points increases since point clouds can be generated with overlapping.
This decreases the number of holes in depth images, but it takes a long time for the rendering process.

Throughout the paper, we use a subset of training views for point cloud generation and refinement.

\noindent{\bfseries Difference with multi-view stereo (MVS) in point cloud generation.}
MVS is a standard approach for generating a cloud or mesh, from a set of images captured from many different views.
We observed that the proposed point cloud generation method can generate more points than the standard MVS method \cite{openmvs} for similar computational time\footnote{With a standard graphics processing unit (GPU),} the processing time of standard MVS is $86.78$ seconds (sec.) and that of the proposed point cloud generation is $87.12$ sec., both with $20$ views..
This leads to the consequence that a point cloud generated by the proposed method above can improve rendering quality compared to that generated by MVS.
Within the proposed NeRF framework, 
a point cloud generated by the proposed point cloud generation method and that given by the standard MVS method resulted in $31.44$ dB and $30.27$ dB in PSNR, respectively (for the Pavillon dataset \cite{DONeRF}; $\alpha=1/2$, $N=8$).

\subsubsection{Depth estimation from a point cloud in the test stage}

\begin{figure}[t!]
\centering
\includegraphics[width=0.8\linewidth]{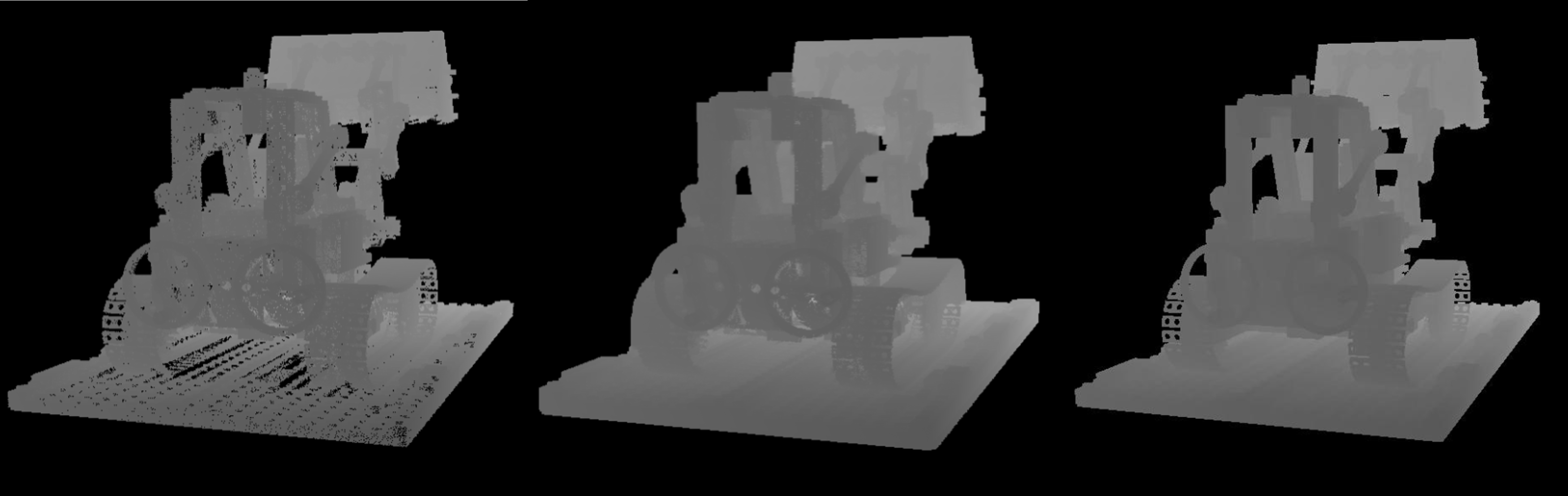}
\caption{A depth image by projecting a point cloud (left);
a depth image by projecting a refined point cloud with hole filling (middle); 
the ground truth depth image (right).}
\label{DepthImg}
\end{figure} 

To obtain a depth image at a novel viewpoint using a point cloud,
we calculate the distance from a 3D point to the camera location by projecting a generated point cloud in Section~\ref{sec:point-gen} to the image plane.
If more than one 3D point are projected onto the same pixel location, we use the closest 3D point to the camera location for distance calculations.

At a novel viewpoint, a projected depth image from a point cloud could have \dquotes{holes}, i.e., pixels with zero values, if those do not have corresponding 3D point(s) in a point cloud.
In projected depth images, however, one cannot identify if such holes correspond to background areas or are missing information on the surface of a foreground object due to limited 3D points.

In this section, we aim to fill-up missing information on the object surface while maintaining background areas.
To distinguish whether holes in projected depth images correspond to background area(s) or missing information on the surface of a foreground object, we use the following condition for a pixel of value $p$:
\begin{equation}
\label{eq4}
\frac{p - \mu}{\sigma} > \kappa,
\end{equation}
where $\mu$ and $\sigma$ is the average and the standard deviation calculated from $M \times M$ neighboring pixels in a projected depth image -- whose center is the pixel of $p$ value -- respectively, and $\kappa$ is some specified threshold.
If the condition (\ref{eq4}) is satisfied, we determine that a hole is missing information on the surface, and fill that hole by applying the moving average filter with a kernel of size $M \times M$.
If $\kappa$ is too large, there still may exist many holes with missing information on the surface of an object (not in background area(s)) even after the hole filling process.
If $\kappa$ is too small, however, one may even fill holes in background area(s).
Selecting an appropriate $\kappa$ value can generate more accurate/useful depth images by minimizing missing information on the object surface and mitigating hole-filling the background areas.

Figure \ref{DepthImg} shows examples of estimated depth images without and with the proposed hole filling process, and the ground-truth depth image. 
We observed that the proposed hole filling method estimates missing depth information for a foreground object, giving more appropriate depth maps.
However, a few parts of the background that are supposed to have zero values are filled with some non-zero values.
It is suboptimal in the perspective of depth estimation, but it is a simple method that can provide sufficiently useful information for proposed near-surface sampling in Section~\ref{sec:surface-sample}.

\section{Results and discussion}

\subsection{Datasets}
\label{sec:dataset}

\begin{figure}[t!]
\centering
\includegraphics[width=\linewidth]{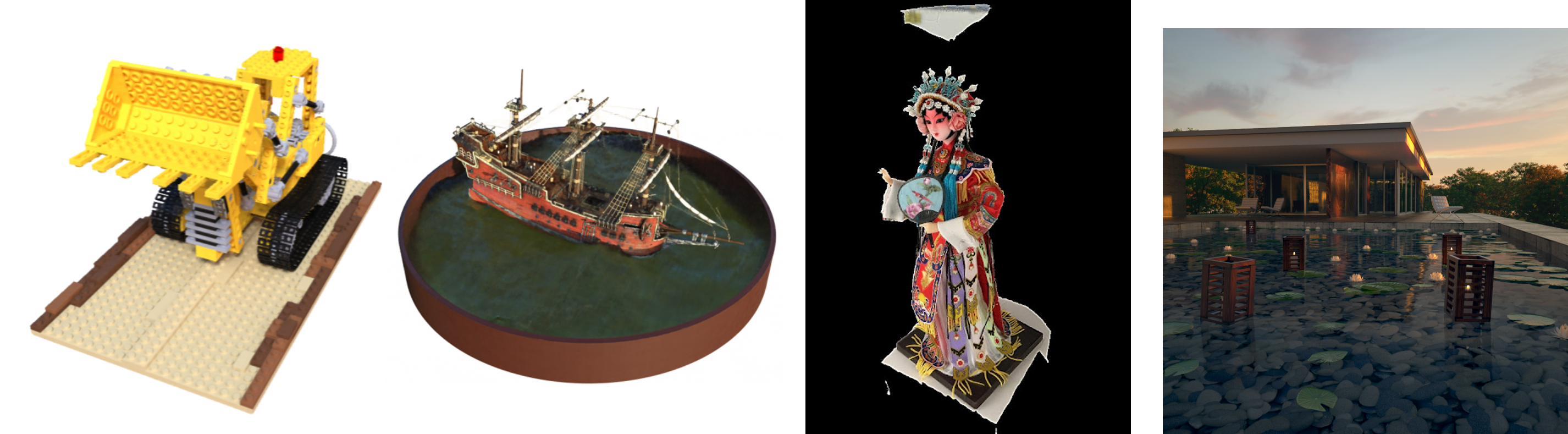}
\caption{The Lego ($1$st), Ship ($2$nd), BlendedMVS ($3$rd), and Pavillon ($4$th) datasets.
}
\label{dataset}
\end{figure} 

We used the synthetic Lego and Ship datasets in original NeRF \cite{NeRF},\footnote{
Each original synthetic dataset consist of $100$ training images and $100$ test images; viewpoints are sampled on the upper hemisphere (with fixed diameter) around an object.}
, the real dataset with the identifier \textsf{5a8aa0fab18050187cbe060e}
 in BlendedMVS \cite{BlendedMVS}, and the Pavillon scene dataset.
Figure \ref{dataset} shows these datasets.
For each synthetic dataset, we used $150$ training images and $50$ test images, all with the spatial resolution of $800 \times 800$.
In generating a point cloud (Section~\ref{sec:point-gen}) for each synthetic dataset, we used $20$ of $100$ training images from the original dataset.\footnote{\label{fn:point-cloud}
We generated a point cloud with $20$ viewpoints by sequentially using the every fifth viewpoint from $100$ viewpoints.
We repeated the point cloud generation process $20$ times, where each iteration consists of four steps in Section~\ref{sec:point-gen}.
(Section~\ref{sec:point-gen} describes the relation between the numbers of viewpoints and repetitions.)}
In constructing a training dataset for each synthetic data,
we selected $50$ of $100$ original test images by skipping one view by one view and added them to the original training dataset.
For the real dataset,
we used $100$ training images and $11$ test images, all with the the resolution of $574 \times 475$.
In generating a point cloud, we used $20$ of $100$ training images.\footnoteref{fn:point-cloud}
For all datasets, each instance has a different viewpoint.
If not further specified, we used the above experimental setup throughout all experiments.

The chosen real data contains multi-view images taken around an object and several images are captured from closer viewpoints to an object.
In our experiments, we used included depth images in \cite{BlendedMVS}, and used blended color images reflecting view-dependent lighting \cite{BlendedMVS}, as the ground truth color images.

We compared the proposed NeRF framework using near-surface sampling with a point cloud, with original NeRF, DONeRF \cite{DONeRF}, DSNeRF \cite{Depth-supervised-NeRF}, and Instant-NGP \cite{instant-ngp}.
For comparing performances between all five methods, we used the re-rendered Lego dataset and Pavillon scene dataset to better fit the view cell methodology of DONeRF that uses additional configurations for view cell generation, and is forward-facing.
We used $210$ training images and $60$ test images, for these comparison experiments.
For a point cloud generation, we used $20$ training images.
For comparing performances between the proposed and original NeRF, we used all the three different datasets (Lego, Ship, and BlendedMVS) that are \emph{not} necessarily forward-facing.

\vspace{-0.3pc}

\subsection{Experimental setup}
\label{sec:experimental setup} 

Throughout experiments with different sampling ranges of the proposed surface-based sampling method, we assumed that the full sampling range of original NeRF \cite{NeRF}, i.e., the radius of the blue fan-shape in Figure~\ref{fig3}, is $4$ (unitless).
For synthetic datasets, we set half of the sampling range of proposed NeRF, i.e., $\alpha$ in (\ref{eq1})--(\ref{eq2}), as $1/2$, $1/4$, $1/8$, and $1/16$.
For real dataset, we set $\alpha$ as $1$, $1/2$, $1/4$, and $1/8$.
(We used larger sampling ranges in real dataset experiments compared to synthetic dataset experiments, since the depth quality of the real dataset is relatively poorer than that of the synthetic dataset.\footnote{
For the \emph{synthetic} Lego and Ship datasets and \emph{real} BlendedMVS dataset, the PSNR value (in dB) for estimated depth in inference is $19.3$, $16.8$, and $10.4$, respectively.})
To see the effects of depth estimation accuracy in the proposed NeRF framework, 
we also ran experiments with ground-truth depth images and estimated depth images via the proposed method.
We set the number of sample points $N=64$, except for experiments using different $N$'s.

In experiments comparing different NeRF methods, we used different numbers of sampling points, i.e., $N$ in (\ref{eq2}).
For fair comparisons, the total number of sampling points per ray of original NeRF is set identical to those of proposed NeRF, DONeRF \cite{DONeRF}, DSNeRF \cite{Depth-supervised-NeRF}, and Instant-NGP \cite{instant-ngp}.
In the original NeRF approach, for each coarse and fine network,
we set the number of sample points per ray to $4$, $8$, $16$, and $32$.
For the proposed NeRF, DONeRF, DSNeRF, we set $N$ as $8$, $16$, $32$, and $64$, and used only one rendering network.
Different from original NeRF that uses samples with different locations for two different networks, 
Instant-NGP uses two networks that estimate color and density respectively, but use samples with the same locations. 
For Instant-NGP, we set the number of samples per ray to $8$, $16$, $32$, and $64$.
That is, in comparing different NeRF methods, we set the total number of sample points per ray as $8$, $16$, $32$, and $64$ consistently for all the NeRF methods.

The remaining hyperparameters of the proposed NeRF approach are listed as follows. 
In determining sampling locations (\ref{eq2}), we randomly sampled $\gamma$ via the uniform distribution between $0$ and $2\alpha/N$.
In the point cloud refinement condition (\ref{eq3}), we set $\tau$ as $0.1$.
In the hole filling condition (\ref{eq4}), we set $\kappa$ and $M$ as $2$ and $11$, respectively. 

We used the following hyperparameters throughout all experiments.
We set the total number of training iterations as $400,\!000$, as the training losses tend to converge after $400,\!000$ iterations.
For each iteration, we set the batch size of input rays as $1024$. 
We used the learning rate of $5 \times 10^{-4}$ until $250,\!000$ iterations, and reduced it to $5 \times 10^{-5}$ after $250,\!000$ iterations.
We used the ADAM optimizer. 

For quantitave comparisons, we used the most representative measure, PSNR in dB, excluding the background area (if available).
We used an NVIDIA GeForce RTX 4090 GPU with 24GB GDDR6X VRAM and 2.31GHz, Intel(R) Xeon(R) Gold 6326 CPU with 2.90GHz, and main memory of 503GB RAM.

\subsection{Comparisons with different sampling ranges in the proposed NeRF framework}
\label{sec:comp:range}

\begin{figure*}[ht!]
\centering
\small\addtolength{\tabcolsep}{-5pt}
\renewcommand{\arraystretch}{1}

\begin{tabular}{cc}
    (a) The Lego dataset & (b) The Ship dataset \\
    \includegraphics[height=3.5cm]{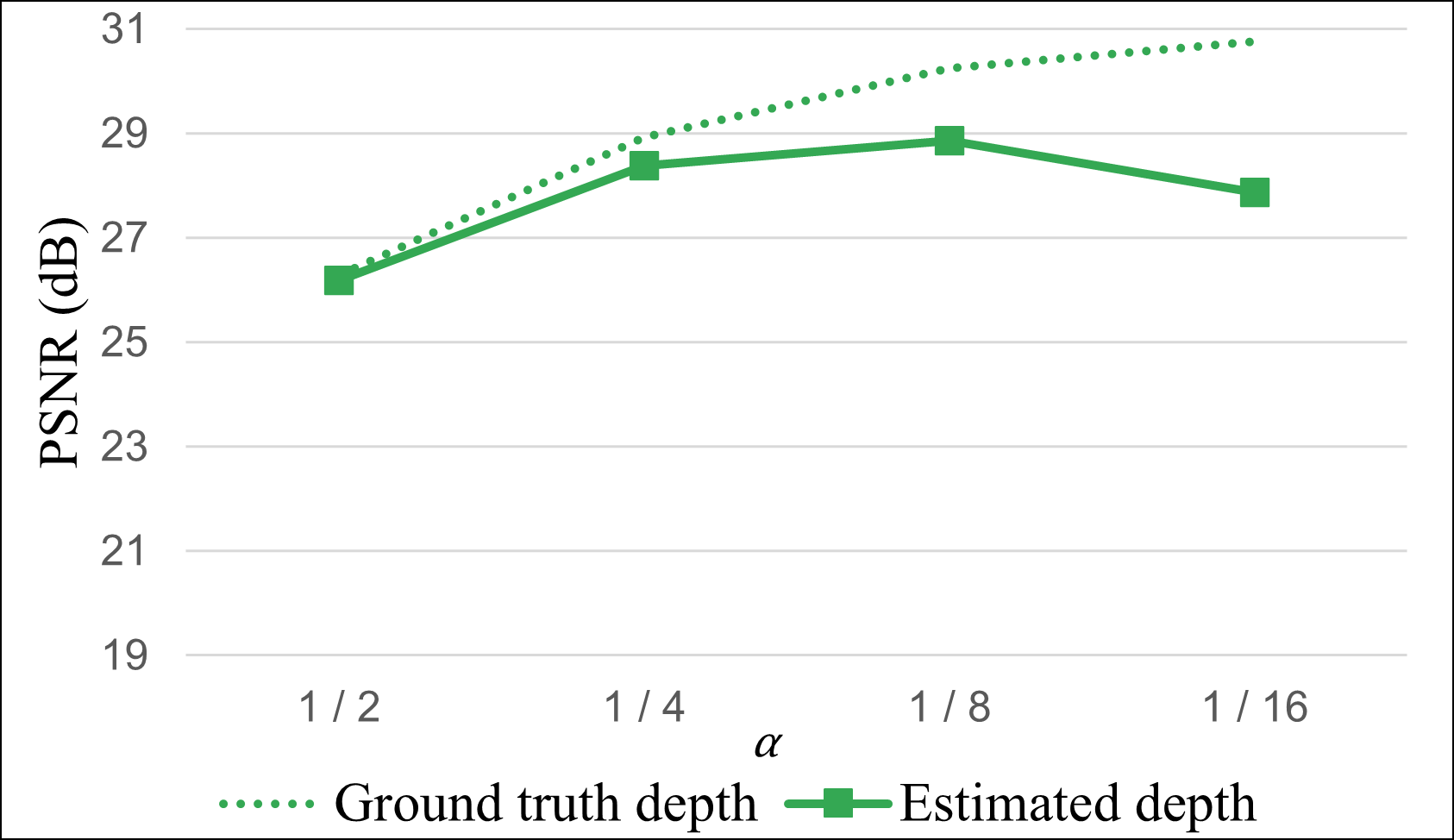} & 
    \includegraphics[height=3.5cm]{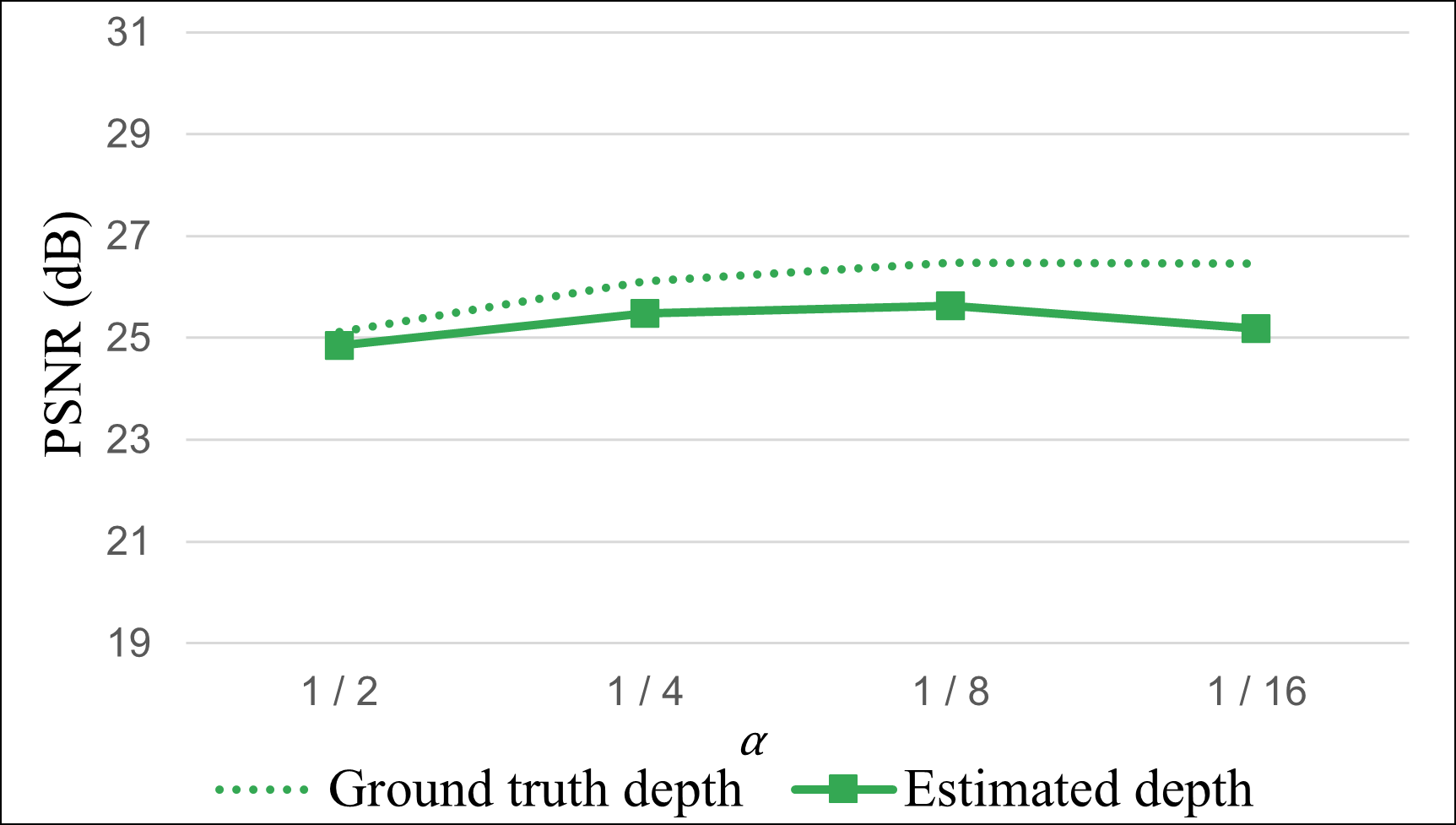} \\
\end{tabular}

\begin{tabular}{c}
    (c) The BlendedMVS dataset \\
    \includegraphics[height=3.5cm]{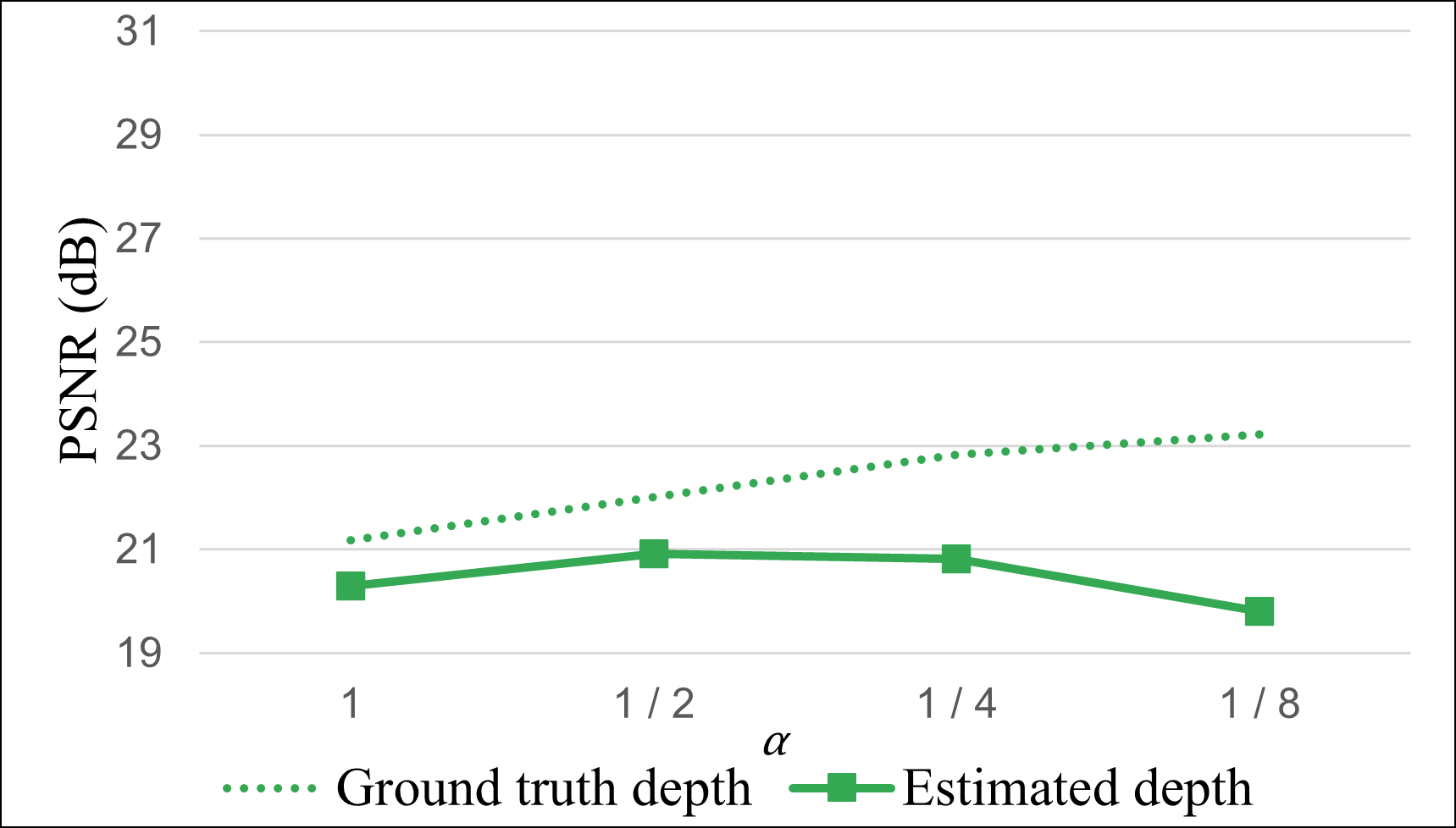} \\
\end{tabular}

\caption{PSNR (dB) comparisons with different sampling ranges, for three different datasets ($N = 64$). 
The dotted and solid lines denote the rendering accuracy in PSNR values of proposed NeRF, with the ground-truth and estimated depth images, respectively.}
\label{PSNR graph per sampling range}
\vspace{-1pc}
\end{figure*}

 \begin{figure*}[ht!]
 \vspace{-1pc}
 \centering
 \small\addtolength{\tabcolsep}{-7.5pt}
 \renewcommand{\arraystretch}{1}

     \begin{tabular}{cccc}
     \multicolumn{4}{c}{} 
     \\
     (a) $\alpha=1/2$ &
     (b) $\alpha=1/4$ &
     (c) $\alpha=1/8$ &
     (d) Ground truth
     \\
     \includegraphics[scale=0.7]{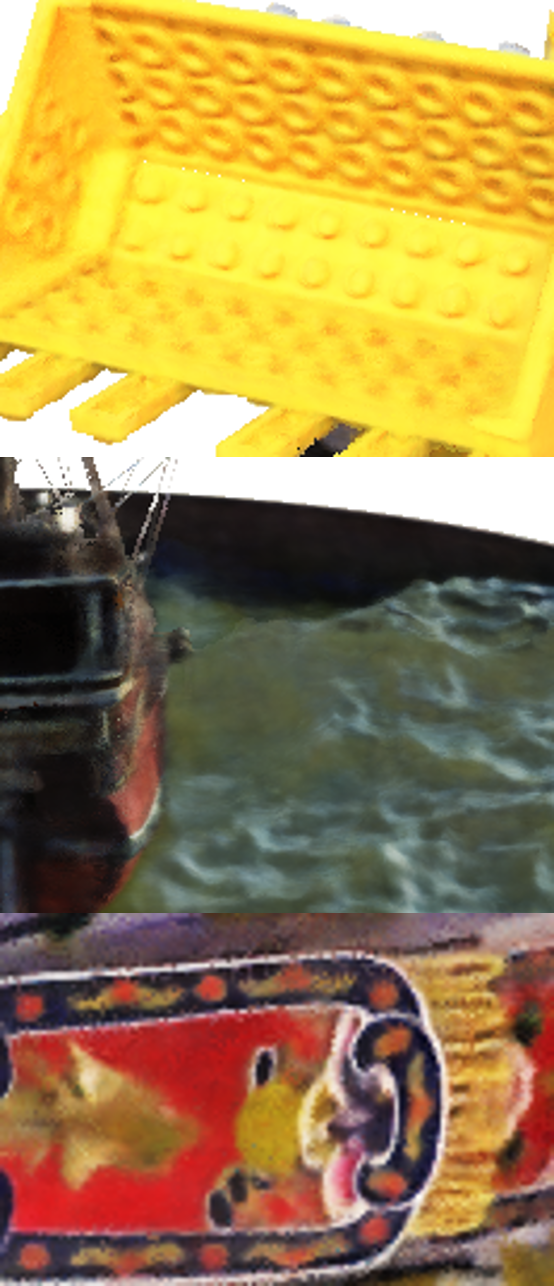}&
     \includegraphics[scale=0.7]{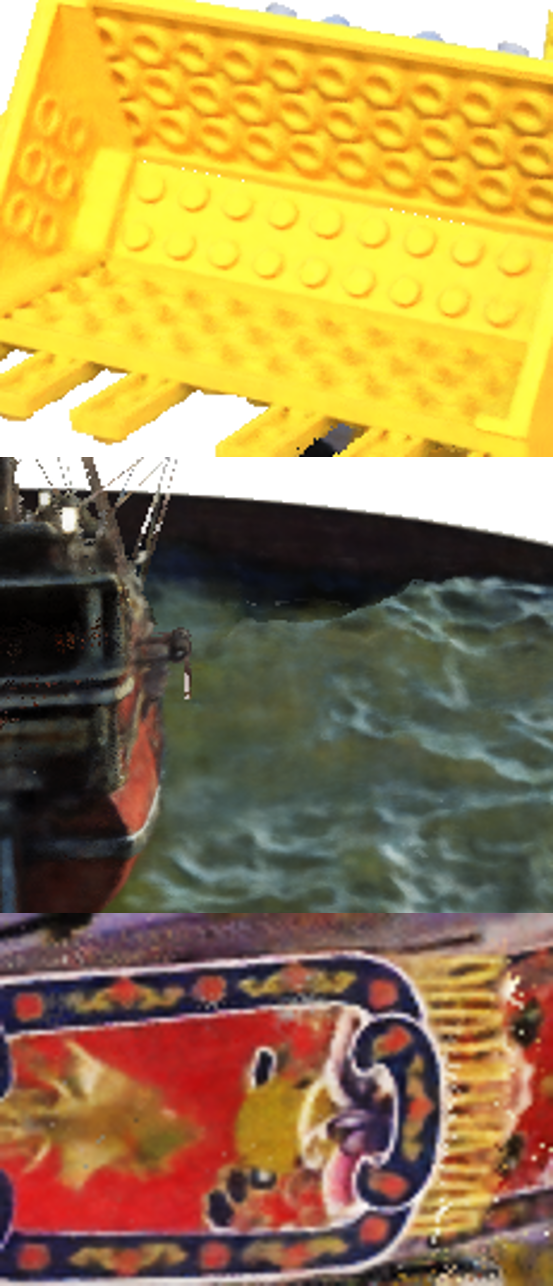} &
     \includegraphics[scale=0.7]{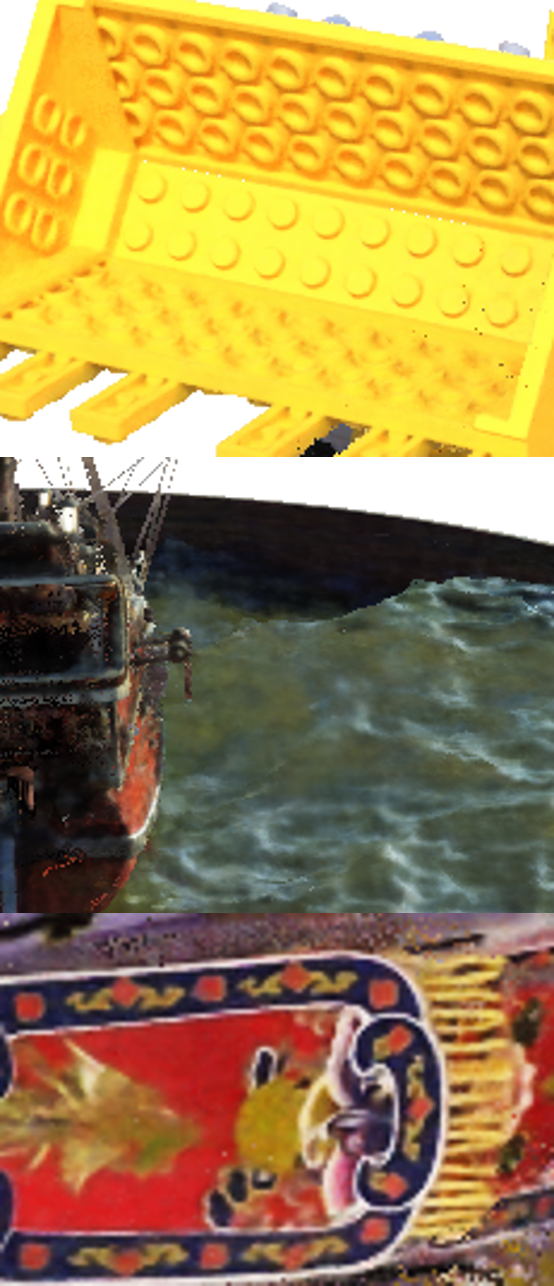} &
     \includegraphics[scale=0.7]{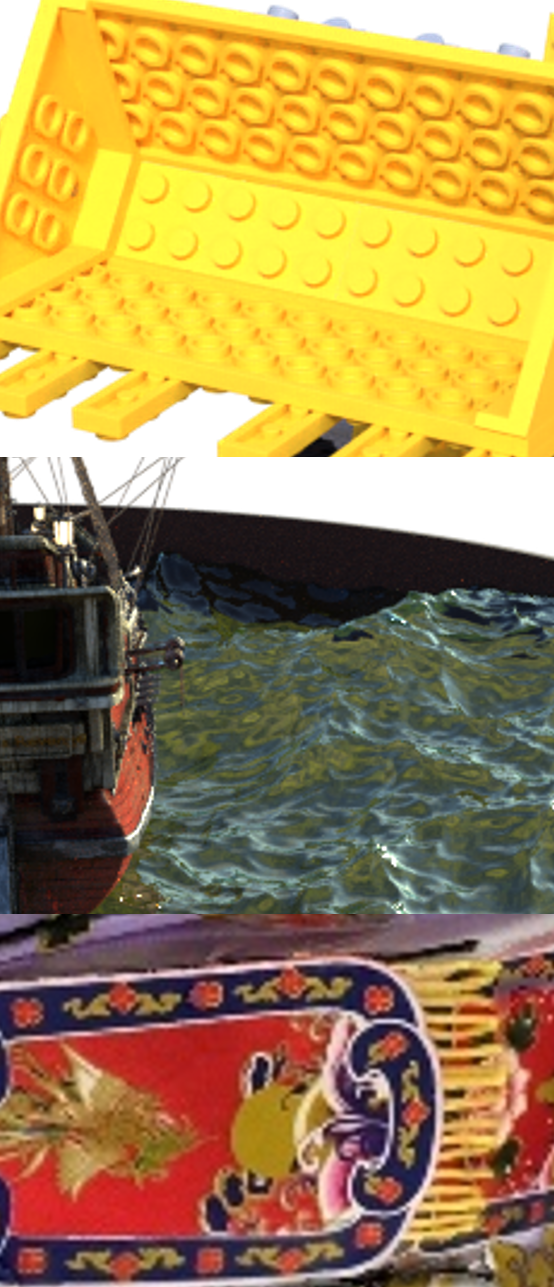}
     \\
     \end{tabular}
 \caption{
Comparisons of rendered images via proposed NeRF for the Lego ($1$st row), Ship ($2$nd row), and BlendedMVS ($3$rd row) datasets, with different sampling ranges (we used estimated depth images via the proposed method; $N = 64$). 
The sampling ranges are scaled versions of the original NeRF's with $\alpha$'s in (\ref{eq2}). 
Images in the $4$th column are ground truths.
 }
 \label{Rendered Img Range}
\vspace{-1.4pc}
 \end{figure*}

Using the proposed surface-based sampling method,
we compared results between different sampling ranges, either with ground-truth or estimated depth images.
First, we compare performances between different sampling ranges, with the ground-truth depth images.
Figure~\ref{PSNR graph per sampling range} with dotted lines compares the rendering quality of proposed NeRF with different sampling ranges, for three different datasets.
It demonstrates that as the sampling range becomes narrow, the rendering quality of NeRF improves.
With the ground truth depth information,
the rendering accuracy improved as the sampling range becomes narrow.
This is natural as the narrower the sampling range, the more sample points are located near the surface of an object.

Next, we compare performances between different sampling ranges, with the estimated depth images via the proposed point cloud generation and hole filling approaches.
Figures~\ref{PSNR graph per sampling range} (solid lines)--\ref{Rendered Img Range} compare the rendering quality of proposed NeRF with different sampling ranges, for three different datasets.
In Figure~\ref{Rendered Img Range}, 
different columns show rendered images with different sampling ranges;
in the last column, the ground truth images are presented;
different rows show rendered images with different datasets.
Figures~\ref{PSNR graph per sampling range}--\ref{Rendered Img Range} demonstrate that the rendering quality of proposed NeRF improves, as the sampling range becomes narrow, but only up to the certain sampling range, e.g., $1/8$ and $1/2$ of the full sampling range of original NeRF for synthetic data and real data, respectively.
If the sampling range is too narrow, 
e.g., $1/16$ and $1/4$ for synthetic data and real data, respectively, the rendering accuracy degraded.
This is because some estimated depth information is inaccurate, but we sample points too near the corresponding inaccurate regions where actual surfaces do not exist.

Finally, we compare the rendering accuracy between two proposed NeRF methods using ground truth and estimated depth images respectively.
Figure~\ref{PSNR graph per sampling range} demonstrates that in the proposed NeRF framework, using estimated depth images degrade the overall rendering accuracy compared to using the ground truth depth, as one may expect.
In particular, points sampled around the inaccurately estimated surface of an object degrade the rendering accuracy.


\subsection{Rendering quality comparisons between different NeRF models}

\subsubsection{Comparisons between five different NeRF models}

\begin{table}[ht!]
\vspace{-2pc}
\centering
\caption{PSNR (dB) comparisons with different numbers of samples per ray for different NeRF methods
($\alpha = 1/16$ and $\alpha = 1/2$ for the Lego and Pavillon datasets in \cite{DONeRF}, respectively).}
\label{tab1}
\begin{tabular}{cccccc}
\specialrule{0.8pt}{1pt}{1pt}
\multicolumn{6}{c}{(a) The Lego dataset}                                                                                            \\ \hline 
\multicolumn{1}{l|}{\diagbox[width=8em]{$N$}{Method}}  & NeRF   & DSNeRF & Instant-NGP  & \multicolumn{1}{c|}{DONeRF}  & Proposed NeRF  \\ \hline
\midrule
\multicolumn{1}{c|}{64}                                & 27.84  & 29.24  & 30.40        & \multicolumn{1}{c|}{31.25}   & \textbf{33.57}           \\
\multicolumn{1}{c|}{32}                                & 26.82  & 28.67  & 30.31        & \multicolumn{1}{c|}{31.13}   & \textbf{32.13}           \\
\multicolumn{1}{c|}{16}                                & 24.58  & 25.81  & 29.13        & \multicolumn{1}{c|}{30.08}   & \textbf{31.36}           \\ 
\multicolumn{1}{c|}{8}                                 & 22.72  & 23.92  & 28.94        & \multicolumn{1}{c|}{29.13}   & \textbf{30.25}          \\ \hline
\multicolumn{6}{c}{(b) The Pavillon scene dataset}                                                                                            \\ \hline 
\multicolumn{1}{l|}{\diagbox[width=8em]{$N$}{Method}}  & NeRF   & DSNeRF & Instant-NGP  & \multicolumn{1}{c|}{DONeRF}  & Proposed NeRF  \\ \hline
\midrule
\multicolumn{1}{c|}{64}                                & 25.13  & 27.61  & 32.09        & \multicolumn{1}{c|}{32.00}   & \textbf{33.21}           \\
\multicolumn{1}{c|}{32}                                & 22.26  & 26.15  & 31.71        & \multicolumn{1}{c|}{31.99}   & \textbf{32.40}           \\
\multicolumn{1}{c|}{16}                                & 19.25  & 23.89  & 30.60        & \multicolumn{1}{c|}{31.71}   & \textbf{31.97}           \\ 
\multicolumn{1}{c|}{8}                                 & 17.52  & 21.18  & 28.42        & \multicolumn{1}{c|}{31.29}   & \textbf{31.44}          \\ \hline
\botrule
\end{tabular}
\vspace{-0.5pc}
\end{table}

 \begin{figure*}[ht!]
 \vspace{-1pc}
 \centering
 \small\addtolength{\tabcolsep}{-7.5pt}
 \renewcommand{\arraystretch}{1}

     \begin{tabular}{c@{\hspace{1mm}}c@{\hspace{1mm}}c}
     \multicolumn{3}{c}{} 
     \\
     (a) Original NeRF&
     (b) DSNeRF &
     (c) Instant-NGP
     \\
     \includegraphics[scale=0.95]{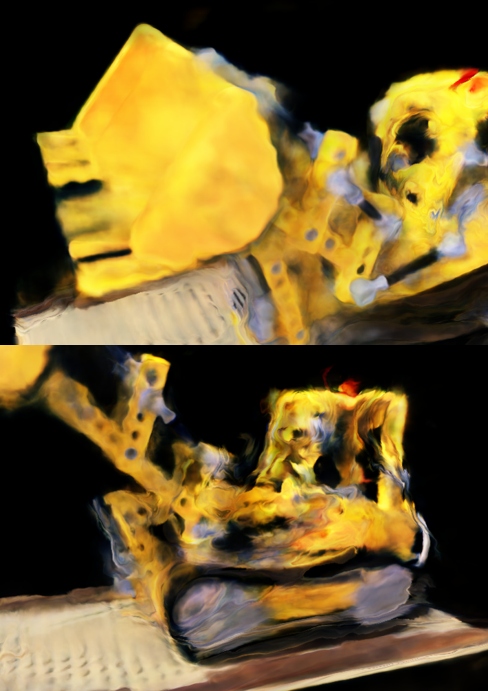}&
     \includegraphics[scale=0.95]{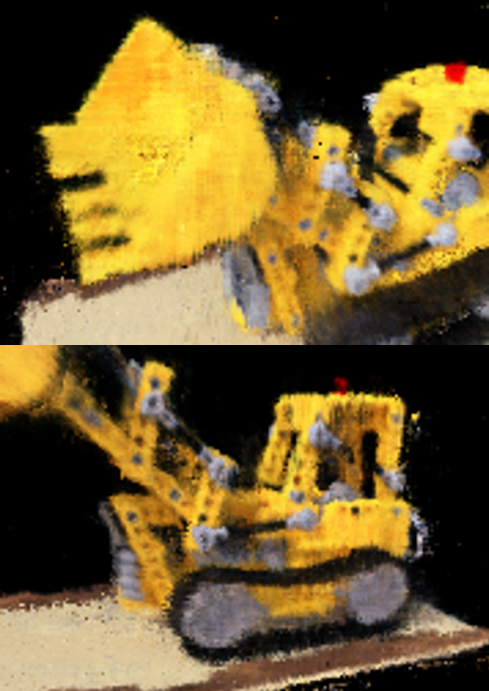} &
     \includegraphics[scale=0.95]{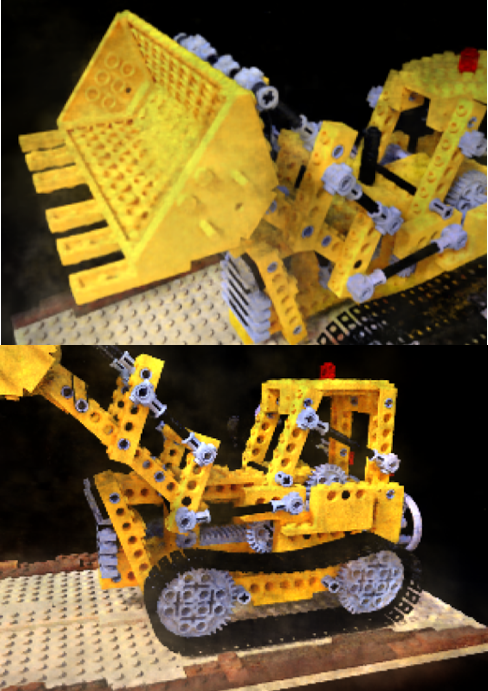} 
     \\
     (d) DONeRF &
     (e) Proposed NeRF &
     (f) Ground truth 
     \\
     \includegraphics[scale=0.95]{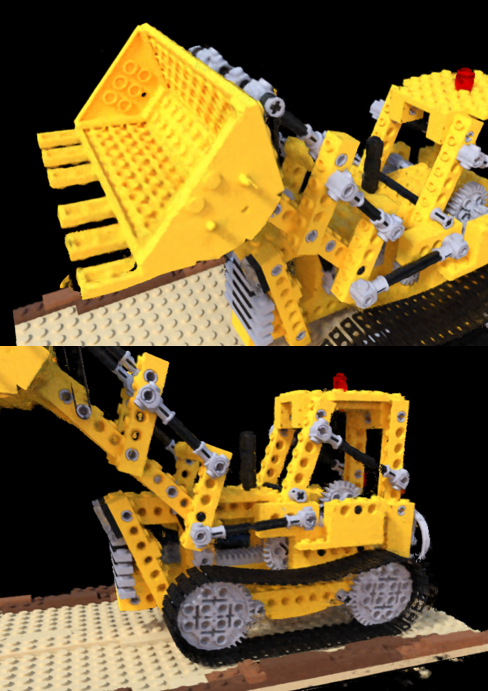} &
     \includegraphics[scale=0.95]{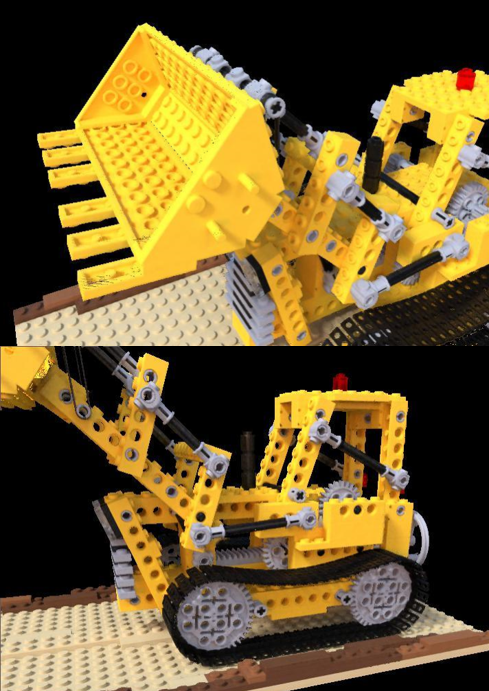} &
     \includegraphics[scale=0.95]{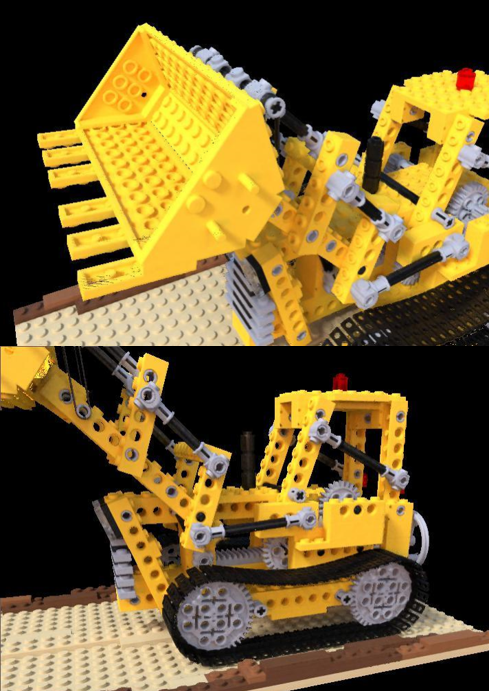}
     \end{tabular}
 \caption{
Comparisons of rendered images with different NeRFs  (the Lego dataset \cite{DONeRF}; $N=8$, $\alpha= 1/16$)
 }
 \label{different nerf models 1}
 \end{figure*}

 \begin{figure*}[ht!]
 \centering
 \small\addtolength{\tabcolsep}{-7.5pt}
 \renewcommand{\arraystretch}{1}

     \begin{tabular}{c@{\hspace{1mm}}c@{\hspace{1mm}}c}
     \multicolumn{3}{c}{} 
     \\
     (a) Original NeRF &
     (b) DSNeRF &
     (c) Instant-NGP 
     \\
     \includegraphics[scale=0.95]{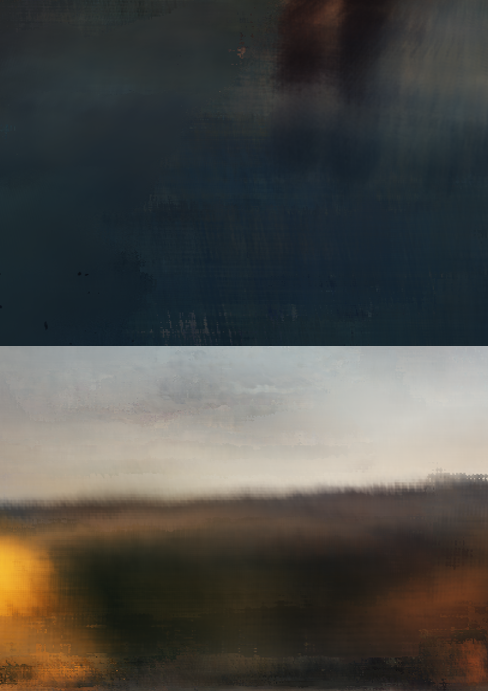}&
     \includegraphics[scale=0.95]{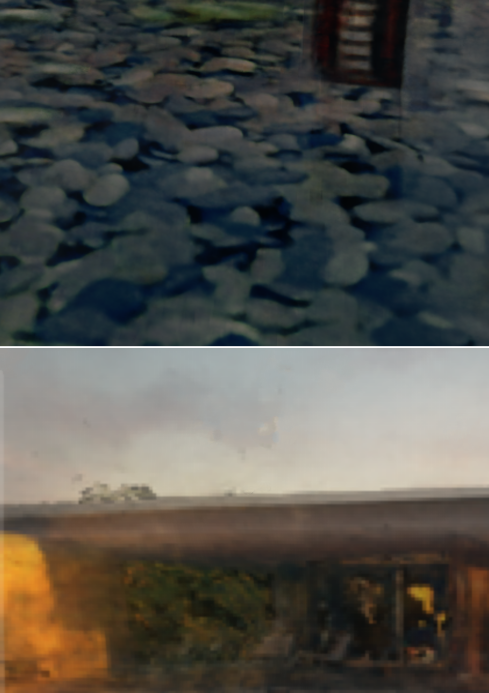}&
     \includegraphics[scale=0.95]{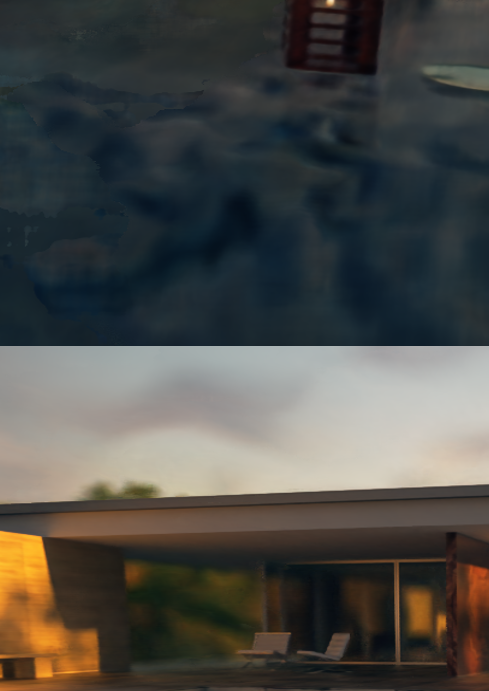} 
     \\
     (d) DONeRF &
     (e) Proposed NeRF &
     (f) Ground truth 
     \\
     \includegraphics[scale=0.95]{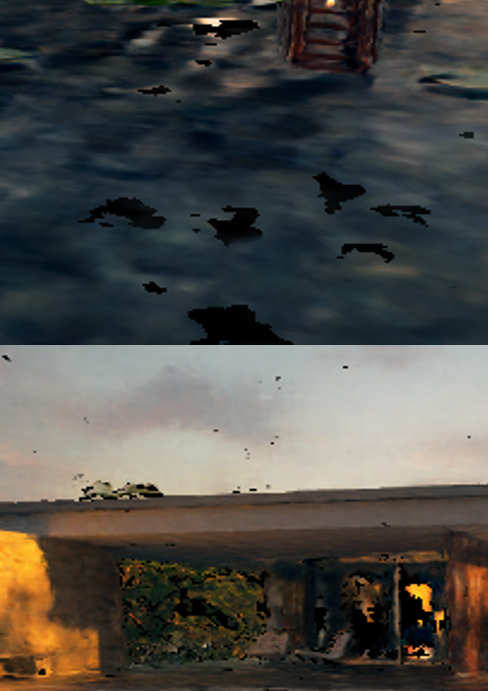}&
     \includegraphics[scale=0.95]{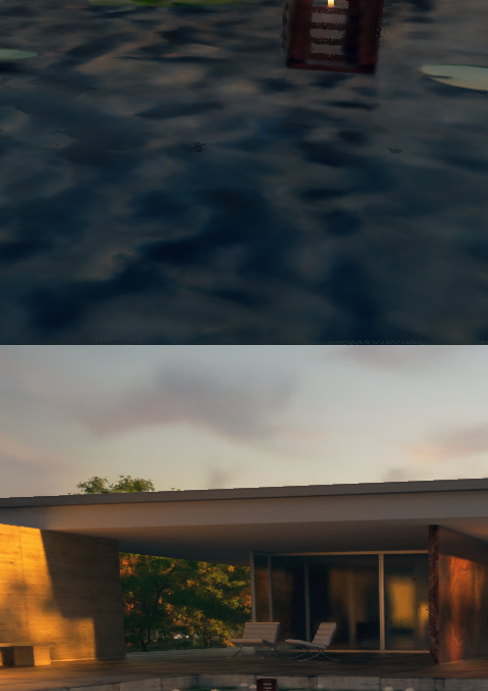}&
     \includegraphics[scale=0.95]{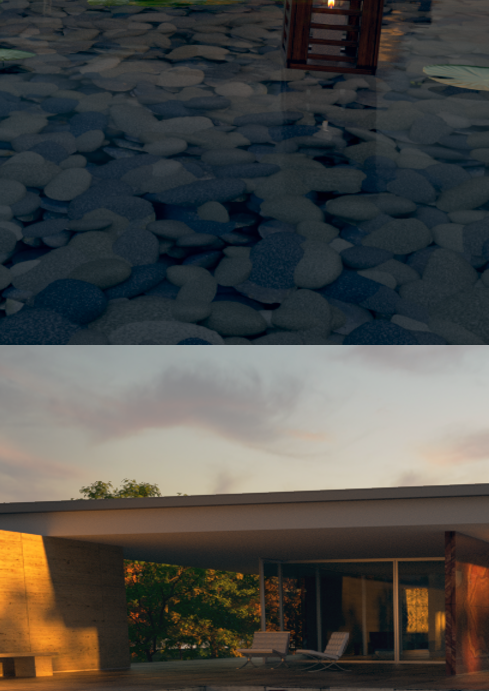}
     \end{tabular}

 \caption{
Comparisons of rendered images with different NeRFs  (the Pavillon scene dataset \cite{DONeRF}; $N=8$, $\alpha= 1/2$)
 }
 \label{different nerf models 2}
\vspace{-1pc}
 \end{figure*}


\begin{figure*}[ht!]
\centering
\small\addtolength{\tabcolsep}{-5pt}
\renewcommand{\arraystretch}{1}

\begin{tabular}{cc}
    (a) The Lego dataset & (b) The Ship dataset \\
    \includegraphics[height=3.5cm]{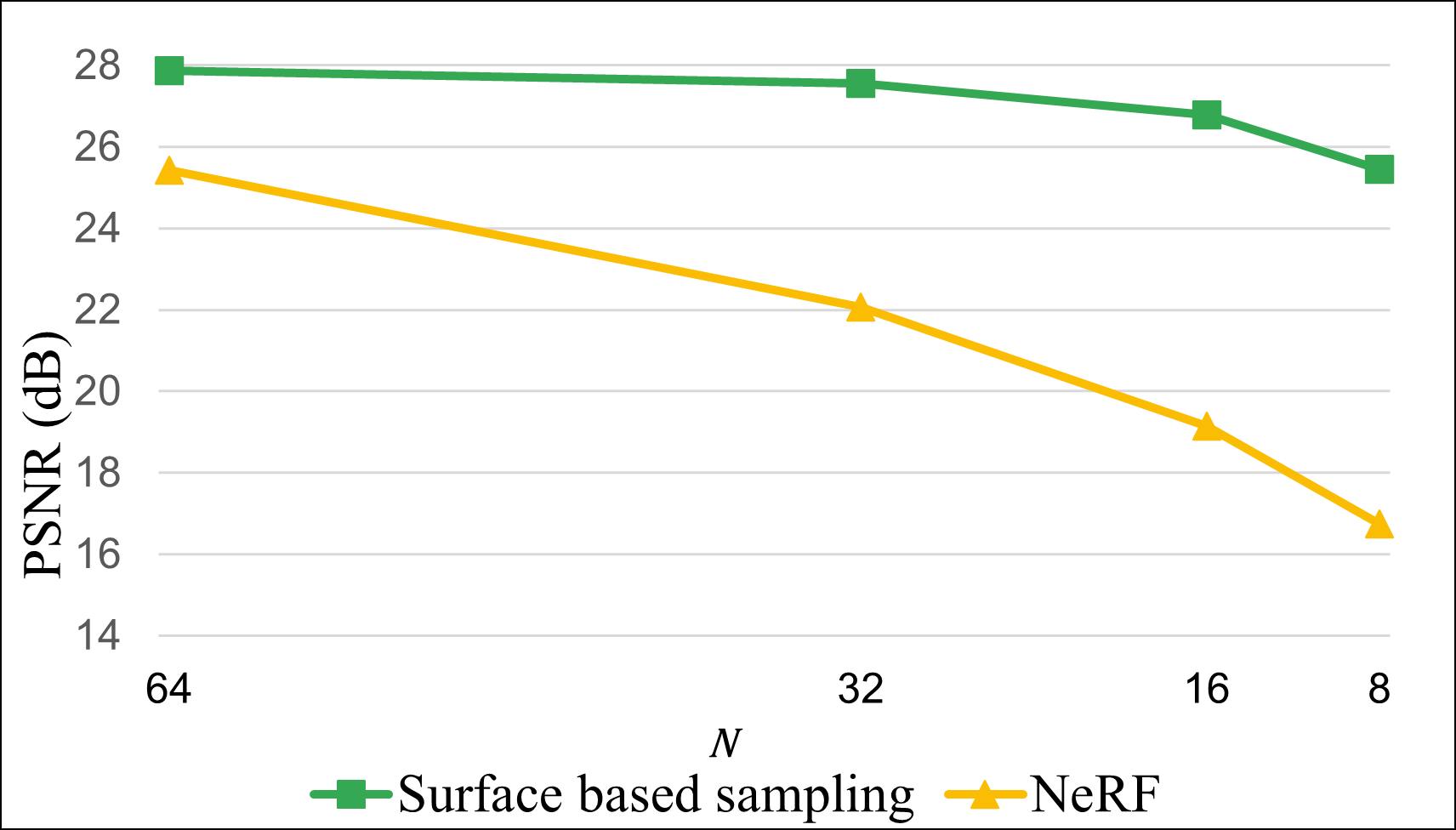} & 
    \includegraphics[height=3.5cm]{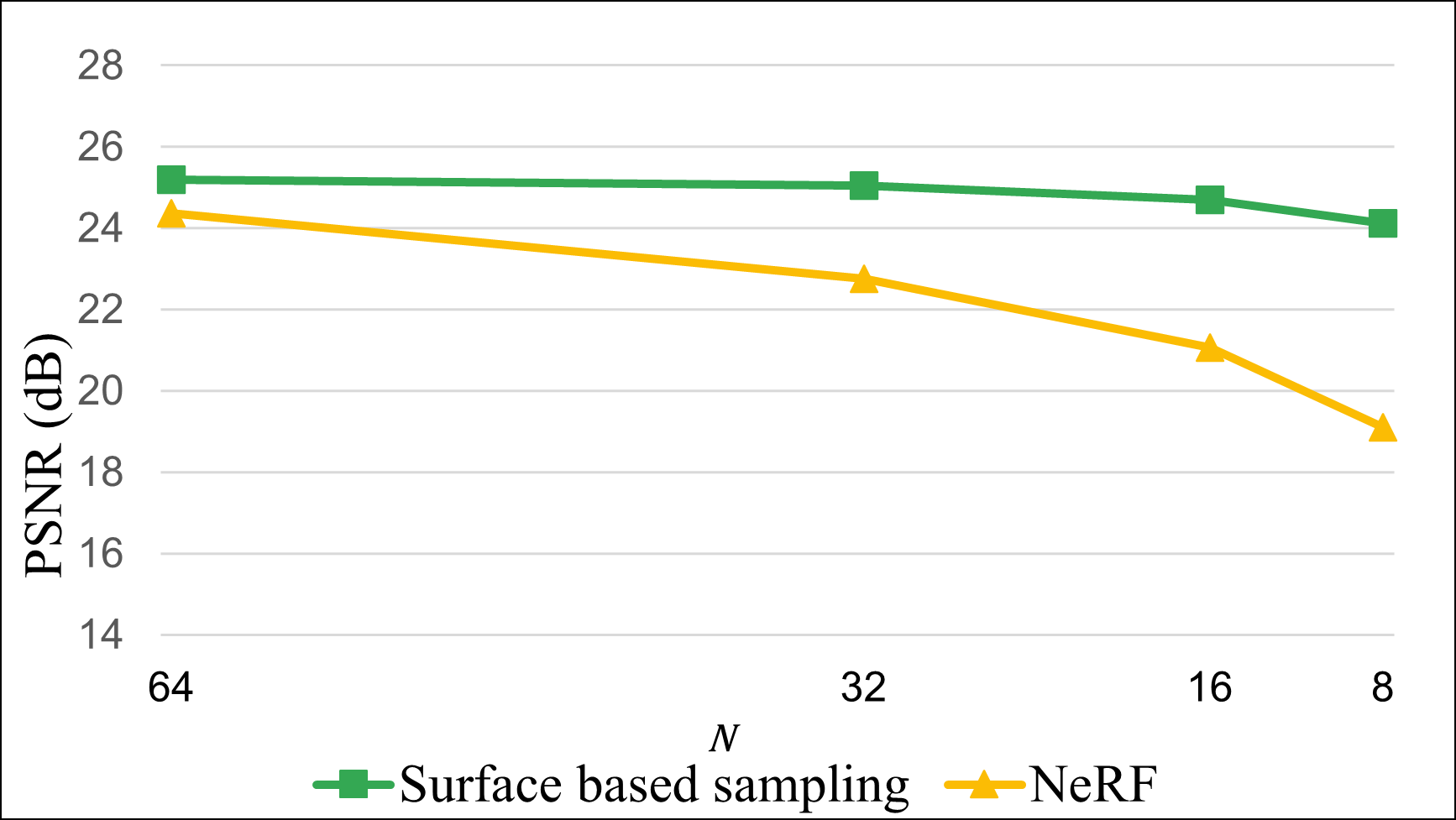} \\
\end{tabular}

\begin{tabular}{c}
    (c) The BlendedMVS dataset \\
    \includegraphics[height=3.5cm]{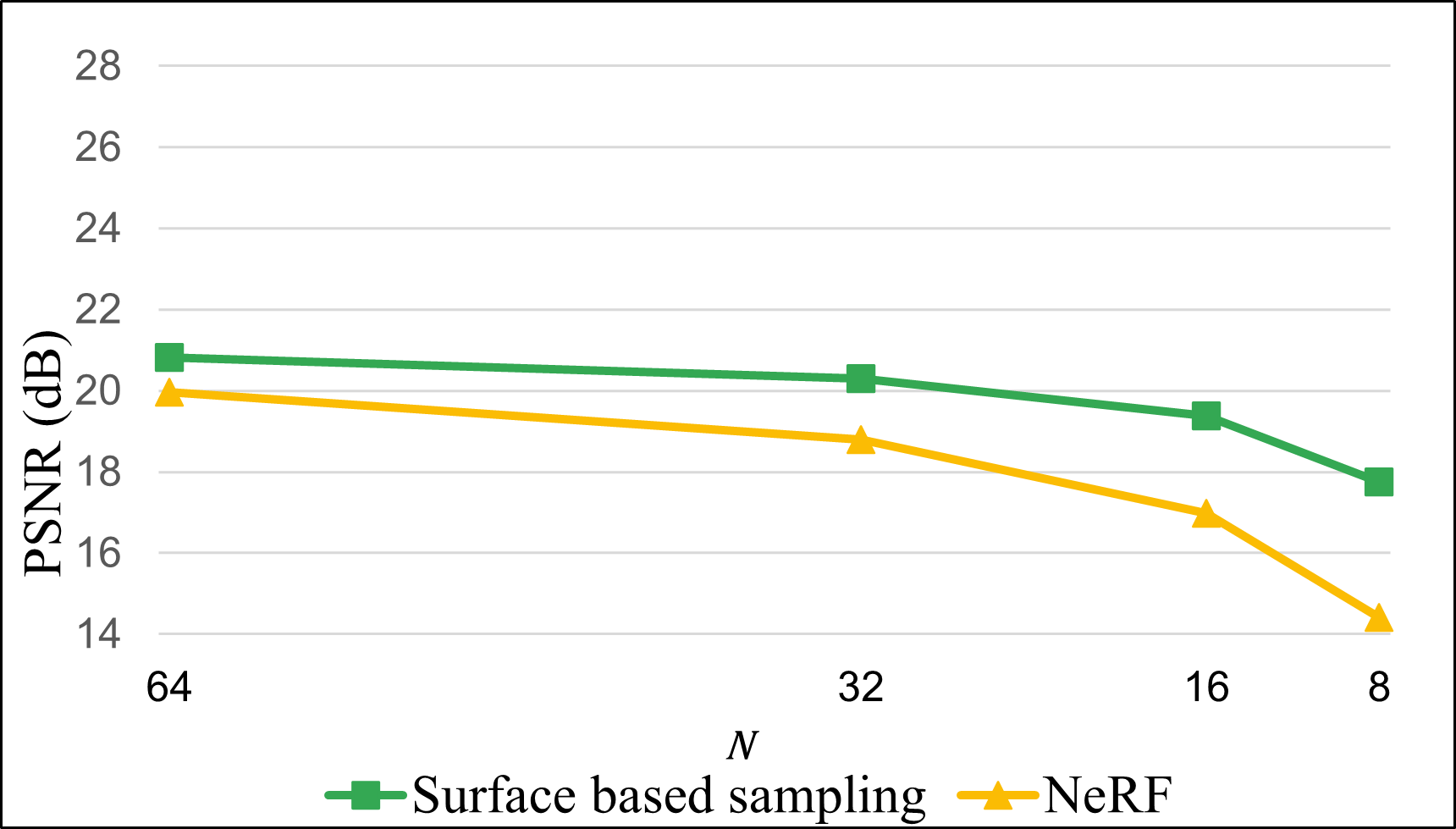} \\
\end{tabular}

\caption{PSNR (dB) comparisons with different numbers of samples per ray, for for three different datasets (for Lego and Ship, $\alpha = 1/16$; for BlendedMVS, $\alpha = 1/4$). 
The green line with squares and yellow line with triangles denote the rendering accuracy of proposed and original NeRF, respectively.}
\label{PSNR graph per number of samples}
\vspace{-1pc}
\end{figure*}

\begin{figure*}[ht!]
 \centering
 \small\addtolength{\tabcolsep}{-7.5pt}
 \renewcommand{\arraystretch}{1}

     \begin{tabular}{ccccc}
     \multicolumn{5}{c}{} 
     \\
     (a) $N=8$ &
     (b) $N=16$ &
     (c) $N=32$ &
     (d) $N=64$ &
     (e) Ground truth
     \\
     \includegraphics[scale=0.63]{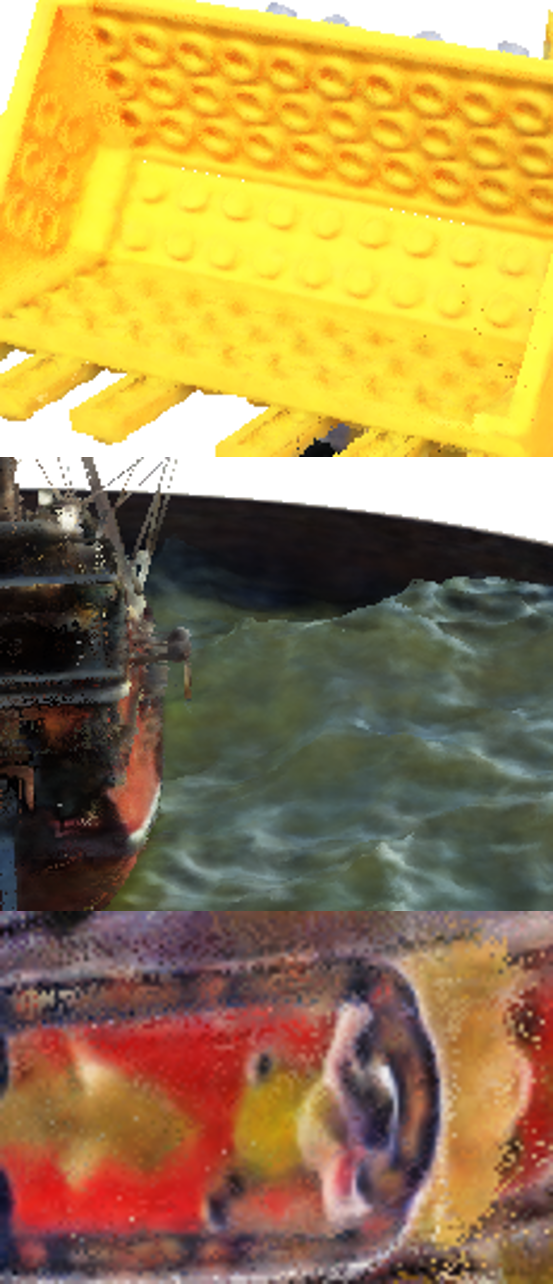}&
     \includegraphics[scale=0.63]{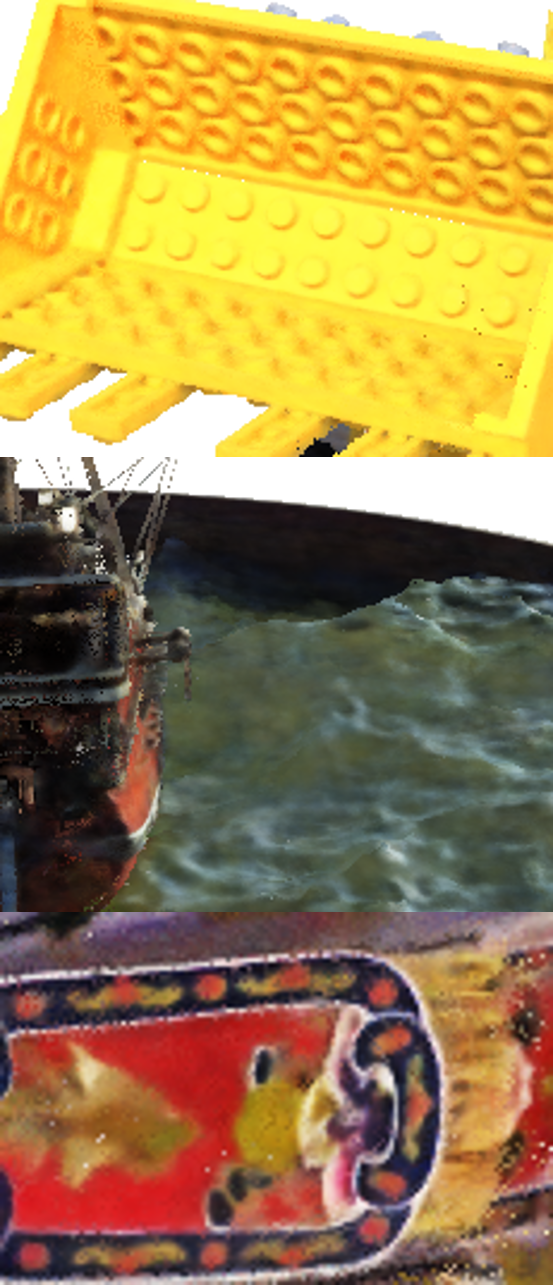} &
     \includegraphics[scale=0.63]{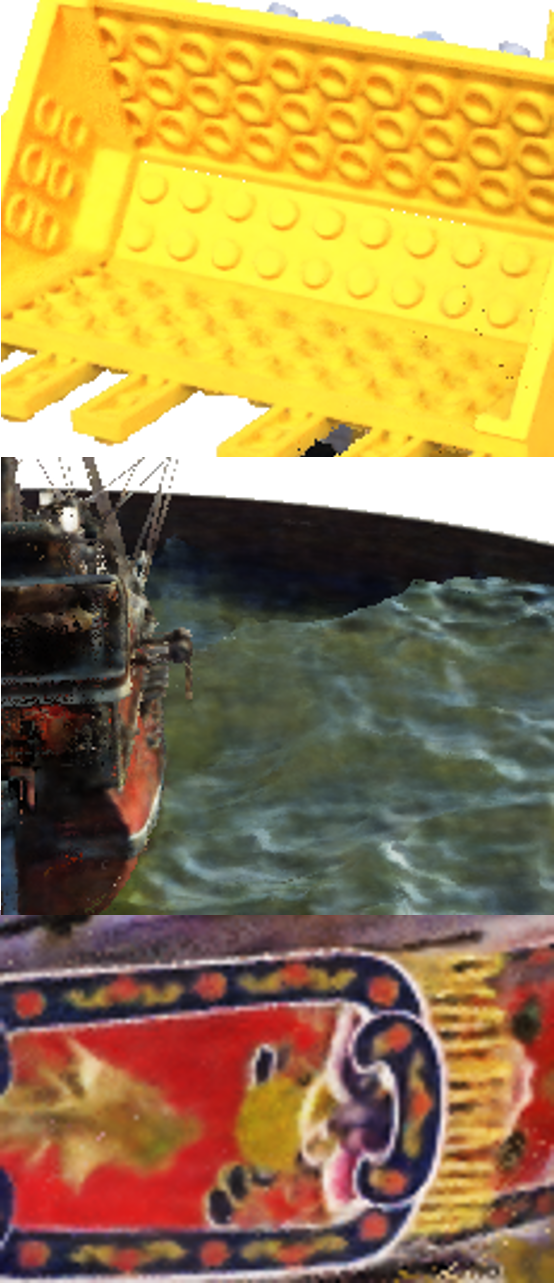} &
     \includegraphics[scale=0.63]{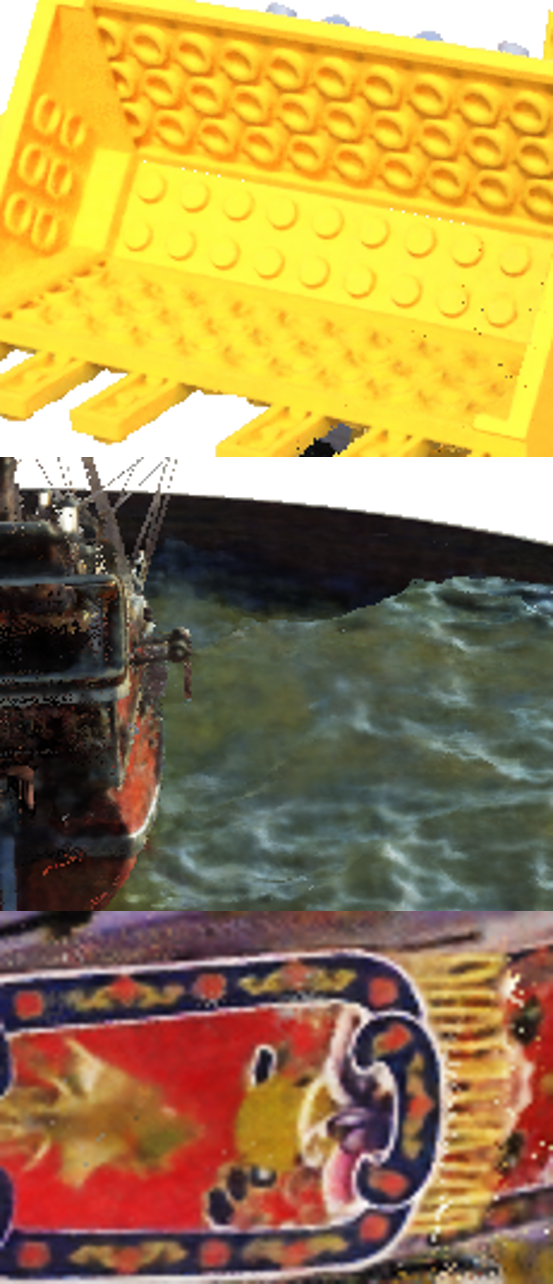} &
     \includegraphics[scale=0.63]{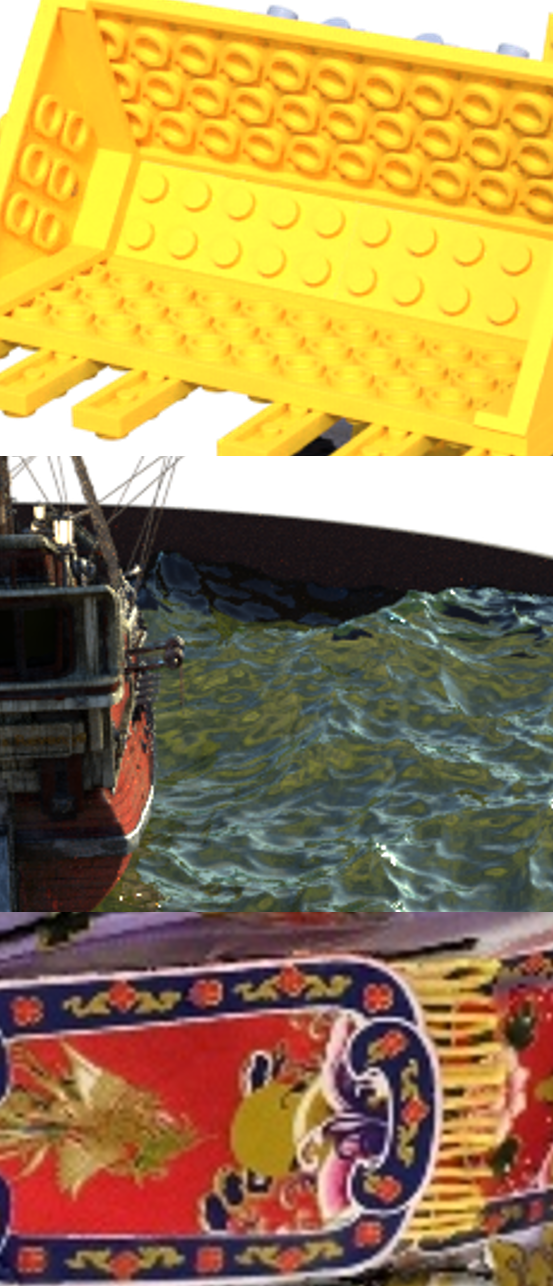}
     \\
     \end{tabular}
 \caption{
Comparisons of rendered images via proposed NeRF for the Lego ($1$st row), Ship ($2$nd row), and BlendedMVS ($3$rd row) datasets, with different numbers of samples per ray (for Lego and Ship, $\alpha = 1/16$; for BlendedMVS, $\alpha = 1/4$). Images in the $5$th column are ground truths.
 }
 \label{Rendered Img Sample}
\vspace{-1.5pc}
 \end{figure*}

\begin{figure*}[ht!]
\centering
 \small\addtolength{\tabcolsep}{-4.5pt}
 \renewcommand{\arraystretch}{1}

     \begin{tabular}{ccc}
    (a) Original NeRF &
    (b) Proposed NeRF &
    (c) Ground truth
    \\
    \includegraphics[width=4cm]{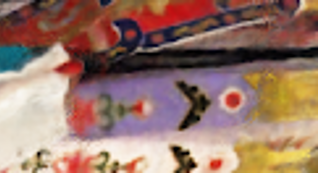} &
    \includegraphics[width=4cm]{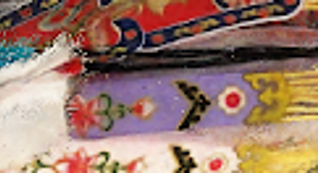} &
    \includegraphics[width=4cm]{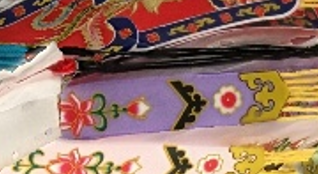}
    \end{tabular}
\caption{A closer look at rendered images by the original NeRF and proposed NeRF method for a real dataset in BlendedMVS \cite{BlendedMVS} ($N = 64$; we used the worst sampling range for the $N = 64$ case, $\alpha = 1/8$). 
}
\label{rendered img comparison with nerf}
\vspace{-1pc}
\end{figure*}

\begin{table}[h!]
\centering
\caption{PSNR (dB) comparisons between the proposed method and original NeRF with a different number of samples and sampling range. The numbers in parentheses denote performance comparisons between the proposed and original NeRF models.}
\label{tab2}
\begin{tabular}[width=0.9\linewidth]{l|ll|lll}
\toprule
$N$                & Method     & Sampling range ($2\alpha$) & Lego                   & Ship                   & BlendedMVS       \\ \hline
\midrule
64                & Original NeRF       & 4               & 25.43                & 24.37                & 19.96          \\ \cmidrule{2-6}
                  & Proposed NeRF & 1           & 26.19 (+0.76)                & 24.85 (+0.48)                & \textbf{20.92} (+0.96) \\
                  &            & 1/2           & 28.38 (+2.95)                & 25.49 (+1.12)                & 20.81 (+0.85)          \\
                  &            & 1/4          & \textbf{28.87} (+3.44)       & \textbf{25.63} (+1.26)       & 19.81 (-0.15)          \\
                  &            & 1/8          & 27.86 (+2.43)                & 25.19 (+0.82)                &                  \\ \Xhline{2\arrayrulewidth}
32                & Original NeRF       & 4               & 22.06                & 22.75                & 18.79          \\ \cmidrule{2-6}
                  & Proposed NeRF & 1           & 23.50 (+1.44)               & 23.57 (+0.82)                & 19.92 (+1.13)          \\
                  &            & 1/2          & 26.05 (+3.99)               & 24.64 (+1.89)               & \textbf{20.29} (+1.50) \\
                  &            & 1/4          & 27.45 (+5.39)               & \textbf{25.09} (+2.34) & 19.54 (+0.75)         \\
                  &            & 1/8          & \textbf{27.55} (+5.49) & 25.04 (+2.29)                &                  \\ \Xhline{2\arrayrulewidth}
16                & Original NeRF       &  4              & 19.15                & 21.07                & 16.99          \\ \cmidrule{2-6}
                  & Proposed NeRF & 1           & 21.10 (+1.95)               & 22.59 (+1.52)               & 18.30 (+1.31)          \\
                  &            & 1/2           & 23.50 (+4.35)                & 23.65 (+2.58)               & \textbf{19.37} (+2.38) \\
                  &            & 1/4          & 25.16 (+6.01)               & 24.23 (+3.16)               & 18.99 (+2.00)         \\
                  &            & 1/8          & \textbf{26.78} (+7.63)       & \textbf{24.70} (+3.63) &                  \\ \Xhline{2\arrayrulewidth}
8                 & Origianl NeRF       & 4               & 16.75                 & 19.11                & 14.42          \\ \cmidrule{2-6}
                  & Proposed NeRF & 1           & 19.51 (+2.76)                & 21.48 (+2.37)               & 16.73 (+2.31)          \\
                  &            & 1/2           & 21.12 (+4.37)               & 22.63 (+3.52)               & 17.74 (+3.32)         \\
                  &            & 1/4          & 22.67 (+5.92)               & 23.21 (+4.10)               & \textbf{18.06} (+3.64) \\
                  &            & 1/8          & \textbf{25.44} (+8.69)       & \textbf{24.11} (+5.00) &                  \\ \hline
\botrule
\end{tabular}%
\end{table}

Table~\ref{tab1} and Figures~\ref{different nerf models 1}--\ref{different nerf models 2} compare the rendering quality between the five different NeRF models, with different number of samples.
The demonstrate that the proposed NeRF outperforms original NeRF, DONeRF, DSNeRF, and Instant-NGP, regardless of the number of sample points per ray.
Figures~\ref{different nerf models 1}--\ref{different nerf models 2} show that the proposed NeRF framework produces significantly better details of a 3D object, compared to the original NeRF, DONeRF, DSNeRF and Instant-NGP.
Table~\ref{tab1} with two different datasets shows that rendering accuracy reduces as the number of sample points per ray decreases.
This is similarly observed in all the five different NeRF models.
This is because as the number of sample point decreases, we have less information to model a 3D object via networks.

\subsubsection{A closer look at original NeRF vs.~proposed NeRF}

Figure~\ref{PSNR graph per number of samples} compares the rendering performance particularly between original and proposed NeRFs, with different numbers of samples per ray.
The figure demonstrates for the three different datasets that the proposed NeRF framework gives significantly better rendering accuracy compared to original NeRF, regardless of the number of sample points per ray.
More importantly, 
Figure~\ref{PSNR graph per number of samples} shows that in the proposed NeRF framework, the performance degradation according to reduction of number of samples per ray is significantly less, compared to original NeRF.
In other words, proposed NeRF can maintain the rendering quality, while reducing the number of samples per ray.
Consequently, we conclude that only with a limited number of samples per ray, the proposed NerF model can achieve significantly better rendering accuracy,
compared to the original NeRF model using many samples per ray.
For the synthetic datasets, the proposed framework using $16$ samples per ray outperformed original NeRF using $64$ samples per ray;
for the real data, the rendering accuracy of the proposed NeRF model using $16$ samples per ray is comparable with that of original NeRF using $64$ samples per ray. 
We expect that the smaller the error in estimated depth at a novel view, the narrower sampling range can be used while reducing the number of samples.

Figure~\ref{Rendered Img Sample} shows rendered images by the proposed framework for different numbers of sample points per ray, with three different datasets.
Except for the extreme case of  using only eight samples per ray ($N=8$), the image quality of rendered images by the proposed framework gradually degraded as the number of samples per ray reduces.
(When $N = 8$, the rendering quality significantly degraded.)
This with the above results from Figure~\ref{PSNR graph per number of samples} underscores the importance of the near-surface sampling approach.

Figure~\ref{rendered img comparison with nerf} compares rendered images by the original and proposed NeRF methods when $N=64$.
Particularly in the proposed NeRF framework, we used the worst sampling range for the BlendedMVS dataset.
The proposed surface-based sampling method significantly improves the overall rendering quality of NeRF, but there exists some dot artifacts.
This is because some missing information still exists or filled holes have inaccurate depth information, after the hole filling.
We conjecture that if one uses a fancier depth estimation method than the proposed simple hole filling scheme, one can remove those artifacts.

Table~\ref{tab2} summarizes PSNR values of the original and proposed NeRF models, for different numbers of samples per ray ($N$) and different sampling range ($2\alpha$).
For each setup using an identical $N$ value,
the proposed NeRF framework outperformed the original NeRF model, regardless of $\alpha$.

\begin{table}[]
\centering
\caption{Training time (hour) comparisons between the proposed method and four different NeRF models with a different number of samples (the Pavillon scene dataset).
We used $400,\!000$ iterations throughout the experiments.}
\label{tab3}
\begin{tabular}{c||cccc|c}
\toprule
\diagbox[width=8em]{$N$}{Method}  & NeRF  & DSNeRF  & Instant-NGP     & DONeRF    & Proposed NeRF  \\ \hline
\midrule
64                                & 21.27  & 16.76  & \textbf{1.54}   & 16.16     & 12.50  \\
32                                & 17.58  & 14.16  & \textbf{0.65}   & 13.66     & 9.34   \\
16                                & 13.27  & 12.44  & \textbf{0.62}   & 11.63     & 7.52   \\
8                                 & 12.14  & 11.86  & \textbf{0.57}   & 11.04     & 7.47   \\ \hline
\botrule
\end{tabular}
\end{table}

\subsection{Training time comparisons between different NeRF models}

Table~\ref{tab3} compares the training time between the five different NeRF methods,
with different numbers of samples.
The Instant-NGP model showed the fastest training time among the five NeRF models {-- note, however, that its rendering accuracy is significantly worse than the proposed NeRF method (see Table~\ref{tab1}).
Except for Instant-NGP, the proposed NeRF method showed the the fastest training time.
Particularly compared to the original NeRF, the proposed NeRF was about two times faster.
The reason is that we trained a single fully-connected network in the proposed NeRF framework, whereas the original NeRF approach trained two fully-connected networks.
It took longer in training DONeRF and DSNeRF than the proposed NeRF model (with the same number of iterations).
This is natural because DONeRF and DSNeRF train an extra depth estimation network.

Regardless of the models, the smaller the number of sample points, it took the less training time.

\section{Conclusion}

In NeRF methods, it is important to reduce the number of sample points per ray while maintaining the rendering quality, as using less samples can reduce training/inference time.
Based on the assumption that the closer the sample point is to the surface of an object, the more important it is for rendering, we propose a near-surface sampling method for NeRF.
The proposed framework samples 3D points only near the surface of an object, by estimating depth images from a 3D point cloud generated with a subset of training data and a simple hole filling method.
For different datasets, the proposed NeRF framework significantly improves the original NeRF \cite{NeRF} and three state-of-the-art NeRF methods, DONeRF \cite{DONeRF}, DSNeRF \cite{Depth-supervised-NeRF}, and Instant-NGP \cite{instant-ngp}.
Particularly compared} to the original NeRF method, 
the proposed framework can achieve significantly better rendering accuracy, with only a quarter of sample points per ray.
In addition, the proposed near-surface sampling framework can accelerate the NeRF training time twice as fast, while improving the rendering quality with an appropriate sampling range parameter.
The proposed method would be useful particularly for applications/technologies where visualizing details is important in novel views.

There are a number of avenues for future work to improve the proposed framework.
First, the proposed framework takes a longer inference time compared to the original NeRF model, because projecting many 3D points to a view plane and estimating a depth image is slower than inference via coarse network in original NeRF.  
We expect to reduce rendering time by speeding up the point cloud projection process. 
Second, the proposed NeRF framework is not completely end-to-end. 
In particular, the point cloud generation and refinement process is in the offline stage and not yet optimized for rendering.
Therefore, we expect to improve the performance of the NeRF model by modifying it with the fully end-to-end approach, incorporating point cloud generation and refinement process into training.
Finally, we expect to further improve the rendering performance of the proposed method by using a more accurate depth estimation method.



\section*{Statements and Declarations}

\bmhead{Funding}
The work of H.~B.~Yoo and I.~Y.~Chun was supported in part by 
NRF grants 2022R1F1A1074546 and RS-2023-00213455 funded by MSIT, and 
the BK21 FOUR Project. 
The work of I.~Y.~Chun was additionally supported in part by IITP grant 2019-0-00421 funded by MSIT,
IBS grant R015-D1,
KIAT grant P0022098 funded by MOTIE, 
the KEIT Technology Innovation program grant 20014967 funded by MOTIE, 
SKKU-SMC and SKKU-KBSMC Future Convergence Research Program grants, and 
SKKU seed grants.
The work of H.~M.~Han and S.~S.~Hwang was supported the NRF grant NRF-2022R1C1C1011084 funded by MSIT.

\bmhead{Competing interests}
The authors declare that they have no conflict of interest.

\bmhead{Ethics approval}
Not applicable

\bmhead{Consent to participate}
Not applicable

\bmhead{Consent for publication}
Not applicable

\bmhead{Availability of data and materials}
The NeRF dataset and BlendedMVS dataset are publicly available at \url{https://paperswithcode.com/dataset/nerf} and \url{https://paperswithcode.com/dataset/blendedmvs}

\bmhead{Code availability}
The code in this study is available from the corresponding author on reasonable request.

\bmhead{Authors' contributions}
Conceptualization, H.~B.~Y., H.~M.~H., S.~S.~H., \& I.~Y.~C.; 
data curation, H.~M.~H.; 
formal analysis, H.~B.~Y. \& I.~Y.~C.; 
funding acquisition, S.~S.~H. \& I.~Y.~C.;
investigation, H.~B.~Y. \& H.~M.~H.; 
methodology, H.~B.~Y., H.~M.~H., S.~S.~H., \& I.~Y.~C.; 
project administration, S.~S.~H. \& I.~Y.~C.; 
resources, I.~Y.~C.; 
software, H.~B.~Y. \& H.~M.~H.; 
supervision, S.~S.~H. \& I.~Y.~C.; 
validation, H.~M.~H., S.~S.~H., \& I.~Y.~C.; 
visualization, H.~M.~H.; 
writing---original draft preparation, H.~B.~Y. \& H.~M.~H.; 
writing---review and editing, I.~Y.~C. 
All authors have read and agreed to the published version of the manuscript.

\bibliography{sn-bibliography}

\end{document}